\documentclass[applsci,article,submit,moreauthors]{Definitions/mdpi}

\usepackage{booktabs}
\usepackage{graphicx}
\usepackage{amssymb}
\usepackage{makecell}
\usepackage{amsmath,amssymb}
\usepackage{algorithm}
\usepackage{algpseudocode}
\usepackage{float}
\preto{\abstractkeywords}{\nolinenumbers}

\floatstyle{ruled}
\restylefloat{algorithm}
\firstpage{1} 
\pubvolume{1}
\issuenum{1}
\articlenumber{0}
\pubyear{2026}
\copyrightyear{2026}
\datereceived{ } 
\daterevised{ } 
\dateaccepted{ } 
\datepublished{ } 

\Title{Agri-MRAG: Parameter-Enhanced Multimodal Retrieval-Augmented Generation for Agricultural Visual Question Answering}

\Author{Fuyong Liu $^{1}$, Xiantao Lei $^{2}$ and Xining Zhang $^{2,}$*}

\AuthorNames{Fuyong Liu, Xiantao Lei and Xining Zhang}

\address{%
$^{1}$ \quad School of Information Science and Engineering, Xinjiang College of Science and Technology, Korla 841000, China; fengyong2122@163.com\\
$^{2}$ \quad School of Computer Science, Wuhan University, Wuhan 430072, China; legend@whu.edu.cn
}

\corres{Correspondence: zhangxn@whu.edu.cn}

\abstract{Multimodal Retrieval-Augmented Generation (MRAG) has shown promise for Agricultural Visual Question Answering (Agri-VQA), but existing methods still face challenges in retrieving relevant knowledge while avoiding redundancy and in integrating retrieved evidence beyond simple prompt concatenation. To address these issues, we propose Agri-MRAG, a multimodal retrieval-augmented framework for agricultural VQA. Specifically, we first introduce a semantic-consistency-guided clustering mechanism that combines multimodal clustering and weighted relevance scoring to select high-confidence knowledge subsets and adaptively determine the retrieval size $k$. We then design a parameterized knowledge injection strategy, where retrieved agricultural knowledge is encoded into LoRA adapters and merged into the feed-forward network (FFN) layers of the MLLM, allowing external knowledge to influence generation through parameter-level adaptation. In addition, a perplexity-guided (PPL-guided) routing mechanism is employed to switch between implicit parametric reasoning and explicit retrieval-assisted generation according to model uncertainty, balancing inference efficiency and answer reliability. For evaluation, we use PlantVillageVQA as a broad agricultural benchmark and construct PlantVQA, a region-specific forestry and fruit crop VQA dataset containing 7,514 images and 15,028 expert-verified question-answer pairs. Experiments show that Agri-MRAG achieves ROUGE-L scores of 41.21 and 46.58 on the two datasets, outperforming mRAG by 11.70 and 4.51 points, and VisRAG by 8.80 and 3.39 points, respectively.}


\begin{document}


\section{Introduction}

With the rapid development of smart agriculture and precision farming, artificial intelligence has been increasingly explored to support agricultural diagnosis, field management, and decision-making~\cite{MLLM1}. In practical agricultural production, tasks such as crop disease diagnosis, nutrient deficiency identification, and growth status assessment require both visual symptom recognition and domain-specific agronomic knowledge. However, farmers and non-expert users often lack timely access to professional guidance, which limits the efficiency and reliability of agricultural decision-making~\cite{MLLM2,MLLM3}. Recently, Multimodal Large Language Models (MLLMs)~\cite{MLLM4,MLLM5,MLLM6} have shown strong capabilities in visual understanding and language generation, providing a promising foundation for agricultural visual question answering (Agri-VQA)~\cite{agrivqa1,agrivqa2,agrivqa3}. By jointly processing crop images and natural language questions, MLLMs have the potential to support automated analysis of crop conditions, disease symptoms, and possible causes~\cite{agrivqa4,agrivqa5}. Nevertheless, Agri-VQA remains challenging because many diagnostic questions require fine-grained discrimination of visually similar symptoms and accurate use of specialized pathological knowledge~\cite{agrivqa1,agrivqa2,MRAG1,MRAG2}.

The knowledge-intensive nature of Agri-VQA makes it difficult for general MLLMs to provide reliable diagnostic responses in agricultural scenarios. Specifically, broad-domain pretraining may not sufficiently cover crop pathology, regional disease manifestations, and agronomic terminology, while subtle differences among visually similar symptoms also require fine-grained visual grounding. As a result, MLLMs may generate inaccurate or ambiguous answers in crop disease diagnosis~\cite{MRAG1,MRAG2}. In response, several studies have focused on developing domain-specific MLLMs fine-tuned on agricultural data. For instance, AgriChat~\cite{agrichat} constructs the AgriMM dataset containing over 600,000 question–answer pairs, and employs a parameter-efficient fine-tuning strategy to inject plant pathology knowledge into the model, thereby significantly improving diagnostic reliability in agricultural scenarios. However, the performance of such methods is often constrained by dataset quality and coverage, resulting in limited generalization capability.

Multimodal Retrieval-Augmented Generation (MRAG)~\cite{MRAG3,MRAG5} provides a feasible way to alleviate this problem by introducing external agricultural evidence. For example, mRAG~\cite{mrag} decomposes the pipeline into three stages: retrieval, reranking, and generation, and adopts a self-reflection unified agent framework; VisRAG~\cite{visrag}, in contrast, performs embedding and retrieval directly on document images using a vision-language model, fully preserving layout and visual information and thereby circumventing the information loss caused by conventional text parsing. However, existing MRAG methods are largely oriented toward general scenarios and still exhibit two limitations in the agricultural domain. First, fixed-size retrieval strategies may retrieve insufficient evidence for complex queries or introduce redundant and noisy information for simple ones, especially when different diseases share similar visual symptoms. Second, most methods integrate retrieved knowledge mainly through prompt-level text concatenation, which weakly couples external agricultural evidence with the model’s internal representations and may cause attention distraction when the retrieved context is long or noisy~\cite{prag,MRAGproblem1}. Although P-RAG~\cite{prag} achieves efficient parameter-level knowledge fusion through document parameterization, it remains confined to the text-only modality and has not been extended to multimodal scenarios. These limitations motivate a MRAG framework that can select reliable evidence according to semantic consistency and incorporate retrieved agricultural knowledge more effectively into the model’s generation process~\cite{MRAGproblem2,MRAGproblem3,MRAGproblem4}.

To address these limitations, we propose Agri-MRAG, a multimodal retrieval-augmented framework that combines cluster-guided evidence selection with parameterized knowledge injection for agricultural visual question answering. The framework is designed to improve both the relevance of retrieved agricultural evidence and the way such evidence is incorporated into the generation process. First, to reduce retrieval noise and contextual redundancy, Agri-MRAG introduces a semantic-consistency-guided clustering mechanism for retrieved image-text candidates~\cite{cluster}. Specifically, the retrieved candidates are grouped in a multimodal embedding space, and each cluster is scored according to its relevance to the input image and question. The framework then selects high-confidence evidence subsets from the most relevant clusters and adaptively determines the retrieval size $k$ according to the query, rather than relying on a fixed number of retrieved samples. Second, to move beyond prompt-level knowledge concatenation, Agri-MRAG extends the parameter-level knowledge fusion approach used in P-RAG for text-only data to multimodal agricultural scenarios, encoding retrieved multimodal agricultural evidence as low-rank adaptation (LoRA)~\cite{lora} parameters and dynamically assembling the corresponding adapters during inference. The assembled adapters are injected into the feed-forward network (FFN) layers of the MLLM, allowing retrieved knowledge to influence generation through parameter-level adaptation. In addition, we employ a perplexity-guided (PPL-guided) routing strategy~\cite{ppl} to estimate generation uncertainty and adjust the use of explicit retrieval evidence during inference. For confident cases, the model can rely mainly on implicit parametric reasoning, while for uncertain queries, explicit retrieval-assisted generation is activated to provide additional fine-grained evidence. Through these designs, Agri-MRAG aims to improve evidence relevance, knowledge utilization, and response reliability in fine-grained Agri-VQA.

To evaluate Agri-MRAG across different agricultural settings, we conduct experiments on both a broad Agri-VQA benchmark and a region-specific dataset. Specifically, PlantVillageVQA is used to assess general agricultural visual question answering performance, while PlantVQA is constructed in this study to focus on forestry and fruit crop pest and disease identification in characteristic economic crops. PlantVQA contains 7,514 images and 15,028 expert-verified question-answer pairs, covering crops such as jujube, fragrant pear, apple, citrus, and grape. By combining evaluation on PlantVillageVQA and PlantVQA, this study examines the effectiveness of Agri-MRAG in both general agricultural scenarios and region-specific crop disease diagnosis.

The main contributions of this paper are summarized as follows:
\begin{itemize}

\item We propose Agri-MRAG, a multimodal retrieval-augmented framework for Agri-VQA that combines semantic-consistency-guided evidence selection with parameterized knowledge injection. By clustering retrieved image-text candidates and selecting high-confidence evidence subsets, Agri-MRAG adaptively determines the retrieval size $k$ and reduces redundant or noisy retrieval.

\item We introduce a LoRA-based parameterized knowledge injection strategy to incorporate retrieved agricultural evidence into the MLLM at the parameter level. Instead of relying only on prompt-level text concatenation, the proposed strategy dynamically assembles retrieval-related adapters and injects them into the FFN layers of the MLLM. A PPL-guided routing mechanism is further introduced to adjust the use of explicit retrieval evidence according to model uncertainty.

\item We construct PlantVQA, a region-specific Agri-VQA dataset for forestry and fruit crop pest and disease identification. The dataset contains 7,514 images and 15,028 expert-verified question-answer pairs, covering characteristic economic crops such as jujube, fragrant pear, apple, citrus, and grape.

\item We evaluate Agri-MRAG on both PlantVillageVQA and PlantVQA. Experimental results show that Agri-MRAG achieves ROUGE-L scores of 41.21 and 46.58 on the two datasets, outperforming the conventional mRAG baseline by 11.70 and 4.51 points, and the VisRAG baseline by 8.80 and 3.39 points, respectively, demonstrating the effectiveness of the proposed retrieval selection and knowledge injection strategies for fine-grained agricultural VQA.

\end{itemize}

\section{Materials and Methods}
\subsection{Dataset Description}
To evaluate Agri-MRAG under different agricultural settings, we use two complementary datasets: the public PlantVillageVQA~\cite{plantvillagevqa} dataset and PlantVQA, a region-specific dataset constructed in this study. PlantVillageVQA serves as a broad Agri-VQA benchmark, while PlantVQA focuses on forestry and fruit crop pest and disease identification in characteristic economic crops. This setting enables the evaluation of Agri-MRAG in both general Agri-VQA scenarios and region-specific crop disease diagnosis.

\subsubsection{PlantVillageVQA Dataset}
PlantVillageVQA is a large-scale Agri-VQA dataset constructed from the PlantVillage image corpus. It contains 55,448 images covering 14 crop species and 38 disease conditions, with 193,609 question-answer pairs. The dataset includes nine semantic categories and three levels of cognitive complexity, enabling evaluation from basic visual perception to more complex agricultural reasoning.

\begin{figure}[H]
\centering
\includegraphics[width=\textwidth]{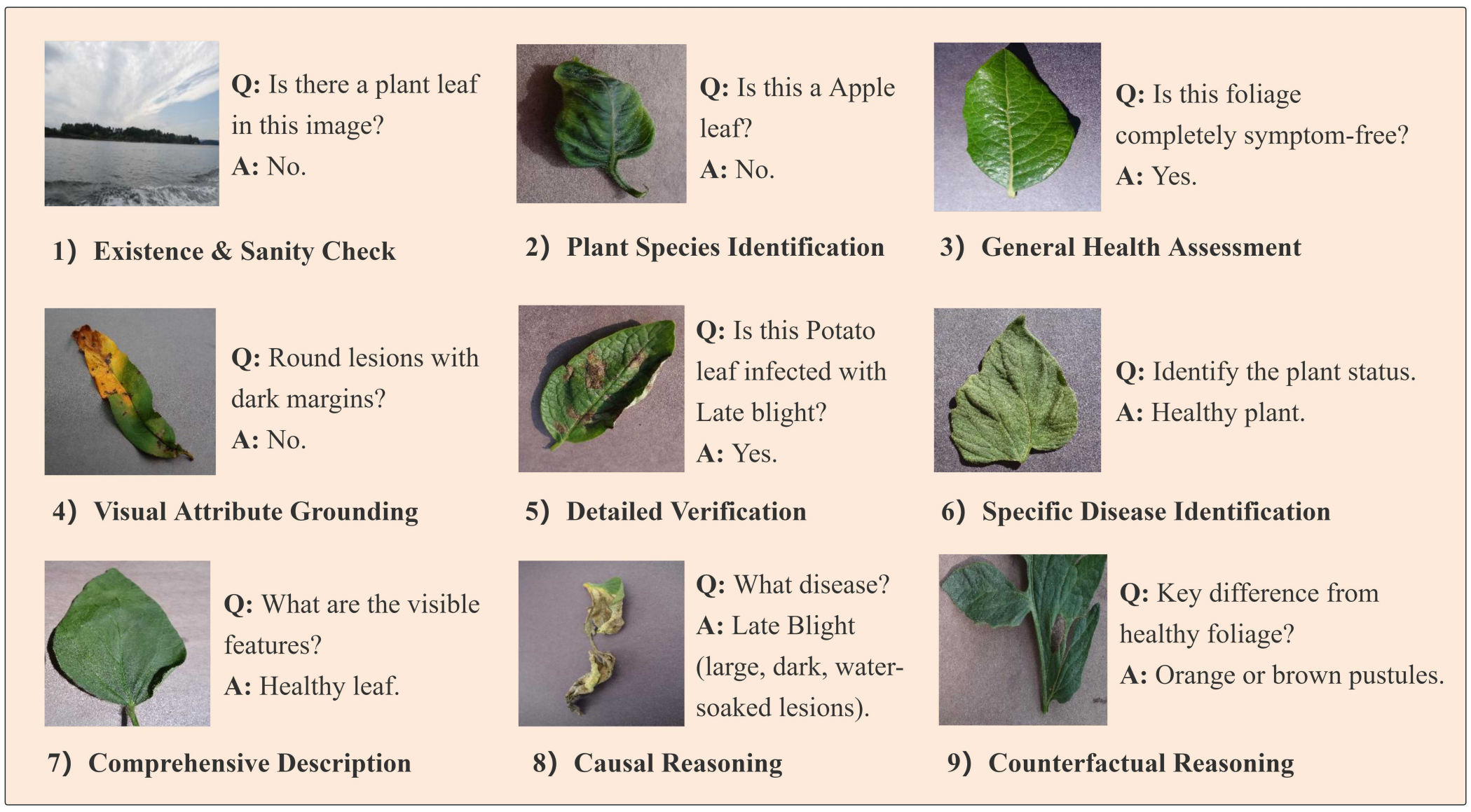}
\caption{Representative images and QA pairs of PlantVillageVQA used in this study.}
\label{fig:plantvillagevqa}
\end{figure}

As shown in Figure~\ref{fig:plantvillagevqa}, PlantVillageVQA covers diverse question types, including existence and sanity check, plant species identification, general health assessment, visual attribute grounding, detailed disease verification, specific disease identification, comprehensive description, causal reasoning, and counterfactual reasoning. These examples show that PlantVillageVQA provides broad coverage of Agri-VQA tasks, especially for leaf-centered disease recognition, plant health assessment, and general agricultural reasoning.

\subsubsection{PlantVQA Dataset}
PlantVQA is a region-specific multimodal VQA dataset constructed in this study for forestry and fruit crop pest and disease identification. It focuses on characteristic economic crops in the Xinjiang Uygur Autonomous Region, China, including jujube, fragrant pear, apple, citrus, and grape. Unlike PlantVillageVQA, which mainly emphasizes general crop disease and leaf-level visual understanding, PlantVQA introduces more forestry and fruit crop scenarios involving fruit surface symptoms, pest damage signs, insect identification, plant part localization, and feature-level judgment.

{During dataset construction, we collected and curated 7,514 representative images of typical diseases and pests affecting forestry and fruit crops. For each image, GLM-4.6V~\cite{glm4.6v} was used to generate detailed symptom descriptions, which were then used as contextual information for generating two visual question-answer pairs. The image descriptions and question–answer pairs were generated before dataset splitting. The generated content was reviewed by agricultural domain experts through duplicate removal, error filtering, and consistency checking against the original agricultural references. Here, duplicate removal refers to the expert regeneration of semantically similar question–answer pairs generated from the same input image, rather than image-level deduplication, so that the diversity and quality of question–answer pairs under the same image can be ensured. The final dataset contains 15,028 question-answer pairs, including 12,038 pairs for training and 2,990 pairs for testing. These pairs correspond to 6,019 distinct images in the training set and 1,495 distinct images in the test set. To ensure no data overlap between the training and test sets, we randomly partitioned the dataset at the individual image level using random seed 2026, to prevent evaluation bias caused by image reuse. No data augmentation was applied during dataset construction, and no image overlap exists between the training and test sets.}

{Beyond the basic statistics of the dataset, we further detail the provenance of the images and the expert verification procedure, which are essential for assessing the reliability of PlantVQA. The images in PlantVQA were collected via web crawling from publicly accessible online sources. Since the images originate from heterogeneous third-party sources, we do not claim a single unified license over the image files; each image record retains its source URL and, where available, the license or terms of use declared at the source. The inclusion criteria were that the image is clearly legible and that the visible symptoms or pest damage are sufficiently clear to support reliable description and question–answer generation.}

{The generated content was reviewed by five agricultural experts, all of whom have worked at the Chinese Academy of Agricultural Sciences (CAAS) for at least three years, with backgrounds in plant pathology, entomology, and plant protection, covering the crop and disease/pest categories included in the dataset. Each generated description and each generated question–answer pair was independently reviewed by all five experts. The review criteria included: (i) factual correctness of the description; (ii) correctness and specificity of the diagnosis or pathogen judgment; (iii) agronomic validity and safety of the management/treatment answers; (iv) consistency between image content and text; and (v) appropriateness of the question type. Disagreements were resolved by majority vote; items without a majority were discussed in subsequent rounds and rejected if still unresolved. Among the initially generated descriptions, approximately 12\% were accepted without modification, 83\% were corrected, and 5\% were rejected. Among the initially generated question–answer pairs, approximately 72\% were accepted without modification, 20\% were corrected, and 8\% were rejected.}

\begin{figure}[H]
\centering
\includegraphics[width=\textwidth]{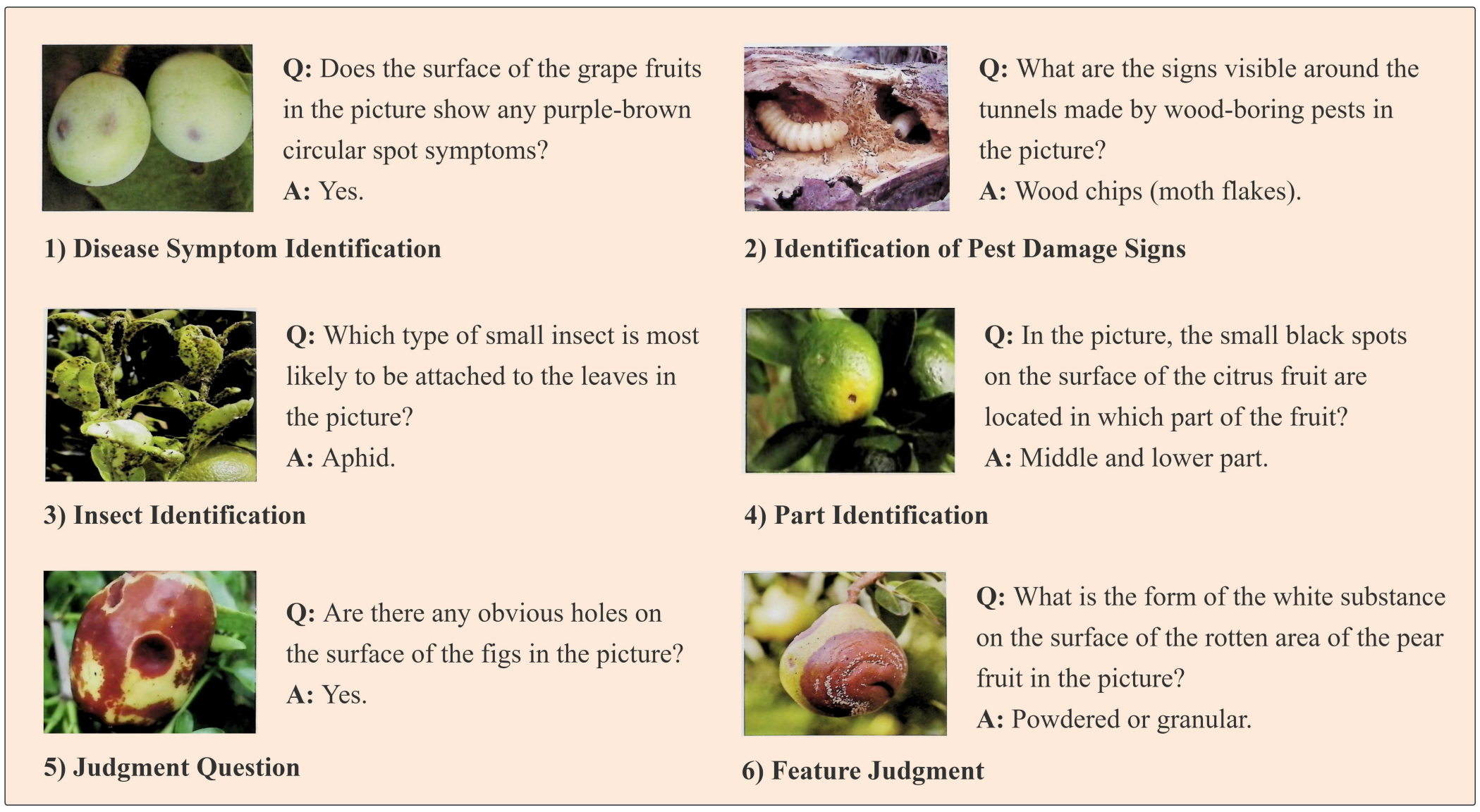}
\caption{Representative images and QA pairs of PlantVQA used in this study.}
\label{fig:plantvqa}
\end{figure}

As illustrated in Figure~\ref{fig:plantvqa}, PlantVQA includes question types such as disease symptom identification, pest damage sign recognition, insect identification, part identification, judgment questions, and feature judgment. These examples highlight the dataset's focus on fine-grained visual evidence in forestry and fruit crop scenarios, such as abnormal fruit-surface patches, wood-boring pest traces, aphid attachment, citrus spot localization, and fruit decay features. Compared with the examples in Figure~\ref{fig:plantvillagevqa}, PlantVQA places greater emphasis on fruit-level symptoms, pest-related visual cues, and region-specific disease knowledge.

PlantVQA provides a complementary evaluation resource for Agri-VQA. Its focus on forestry and fruit crops extends the evaluation beyond general leaf disease recognition and better reflects region-specific agricultural applications. The use of fine-grained symptom descriptions and expert-reviewed question-answer pairs also supports the assessment of models in scenarios requiring detailed visual discrimination and domain-specific agricultural knowledge.
\begin{figure}[H]
\centering
\includegraphics[width=\textwidth]{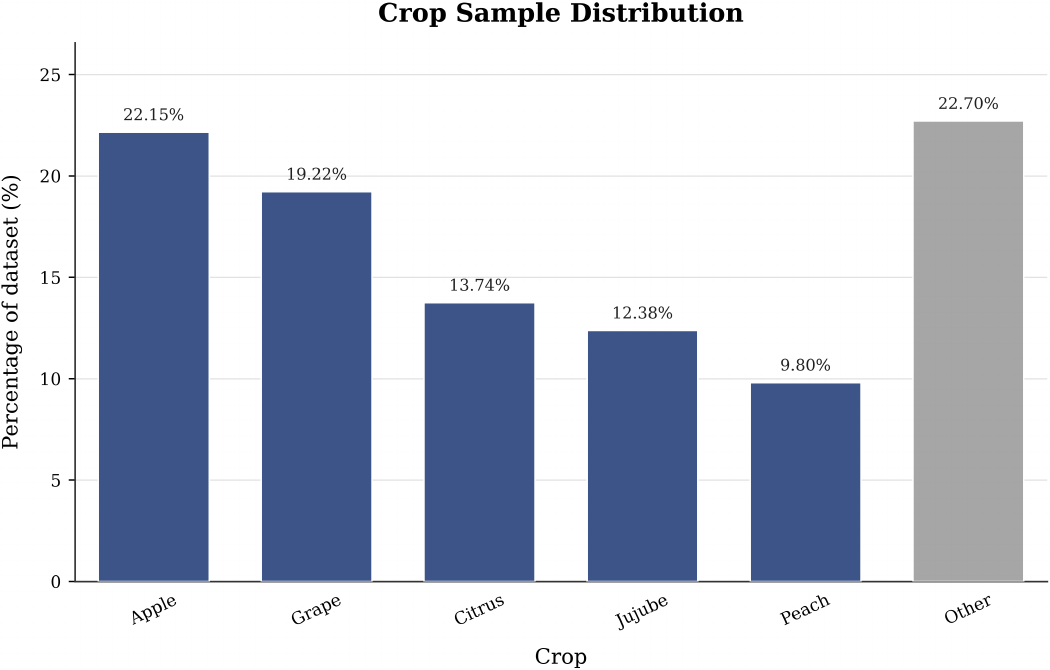}
\caption{{Distribution of Crop Categories in the PlantVQA Dataset.\label{bar1}}}
\end{figure}

\begin{figure}[H]
\centering
\includegraphics[width=\textwidth]{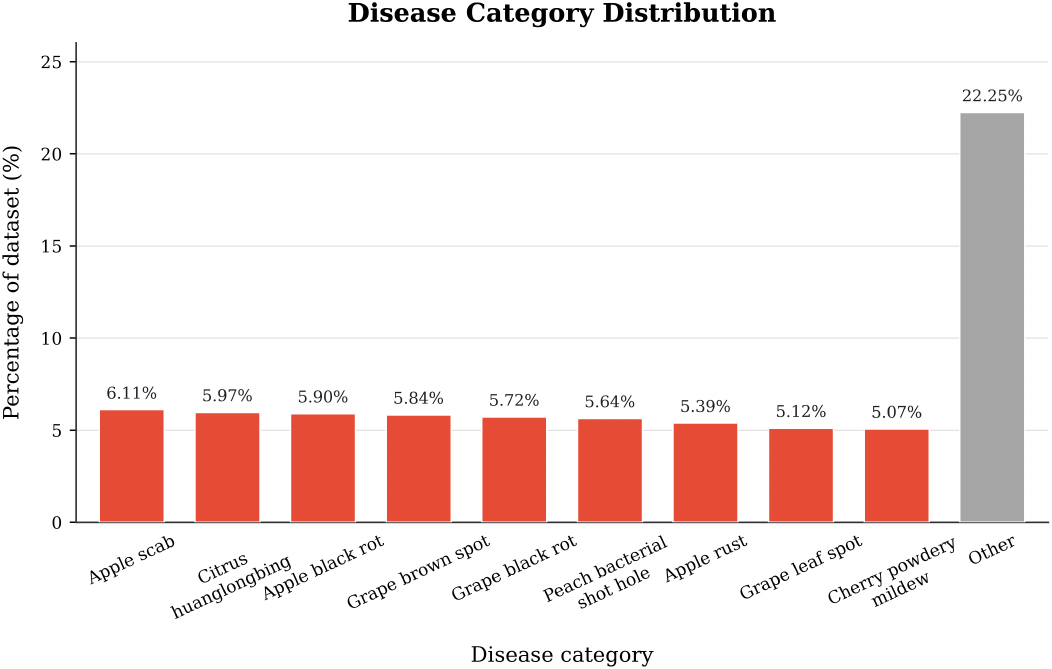}
\caption{{Distribution of Disease Categories in the PlantVQA Dataset.\label{bar2}}}
\end{figure}

\begin{figure}[H]
\centering
\includegraphics[width=\textwidth]{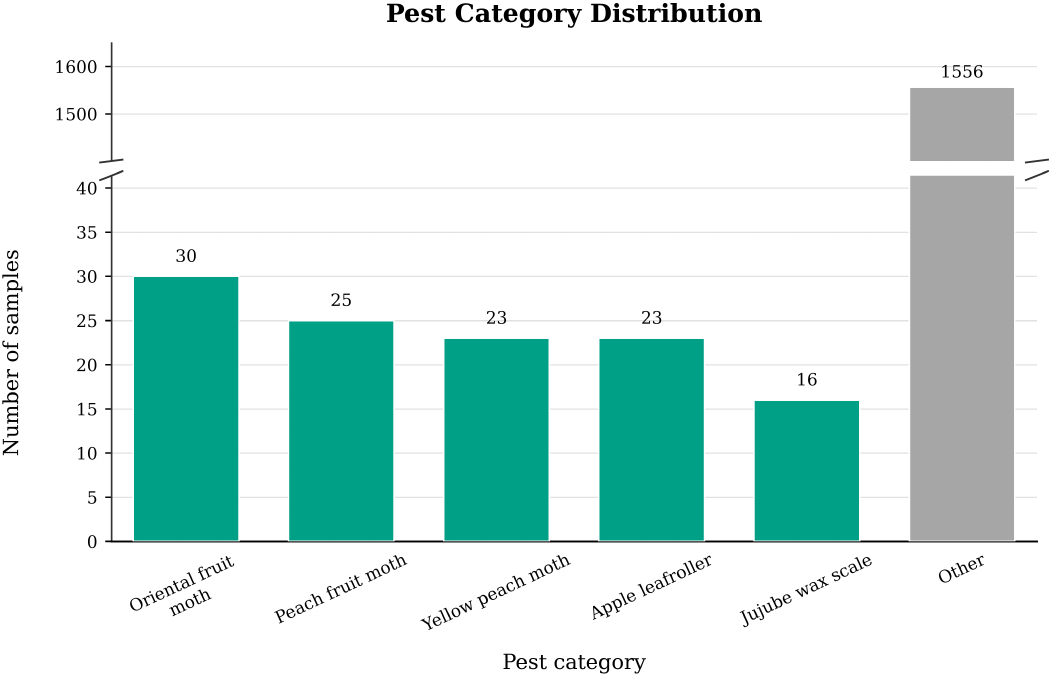}
\caption{{Distribution of Pest Categories in the PlantVQA Dataset\label{bar3}}}
\end{figure}

{For a more comprehensive characterization of PlantVQA, we further analyzed the distributions of crop, disease, and pest categories in the dataset. As shown in Figure \ref{bar1}, Figure \ref{bar2} and Figure \ref{bar3} in terms of crop distribution, apple (22.15\%), grape (19.22\%), citrus (13.74\%), jujube (12.38\%), and peach (9.80\%) are the major categories, while the remaining crops together account for 22.70\%. In terms of disease distribution, apple scab (6.11\%), citrus huanglongbing (5.97\%), apple black rot (5.90\%), and grape brown spot (5.84\%) are among the most frequent categories, with the remaining diseases together accounting for 22.25\%. In terms of pest distribution, oriental fruit moth (30), peach fruit moth (25), yellow peach moth (23), apple leafroller (23), and jujube wax scale (16) are the major pest categories, while the remaining pests together account for 1556 samples. These distributions indicate that PlantVQA provides broad coverage across crop, disease, and pest categories, while also retaining a certain proportion of long-tail categories, which is conducive to evaluating the model's fine-grained recognition capability in region-specific crop scenarios.}
\subsection{Agri-MRAG Framework}
\begin{figure*}[htbp]
\includegraphics[width=\textwidth]{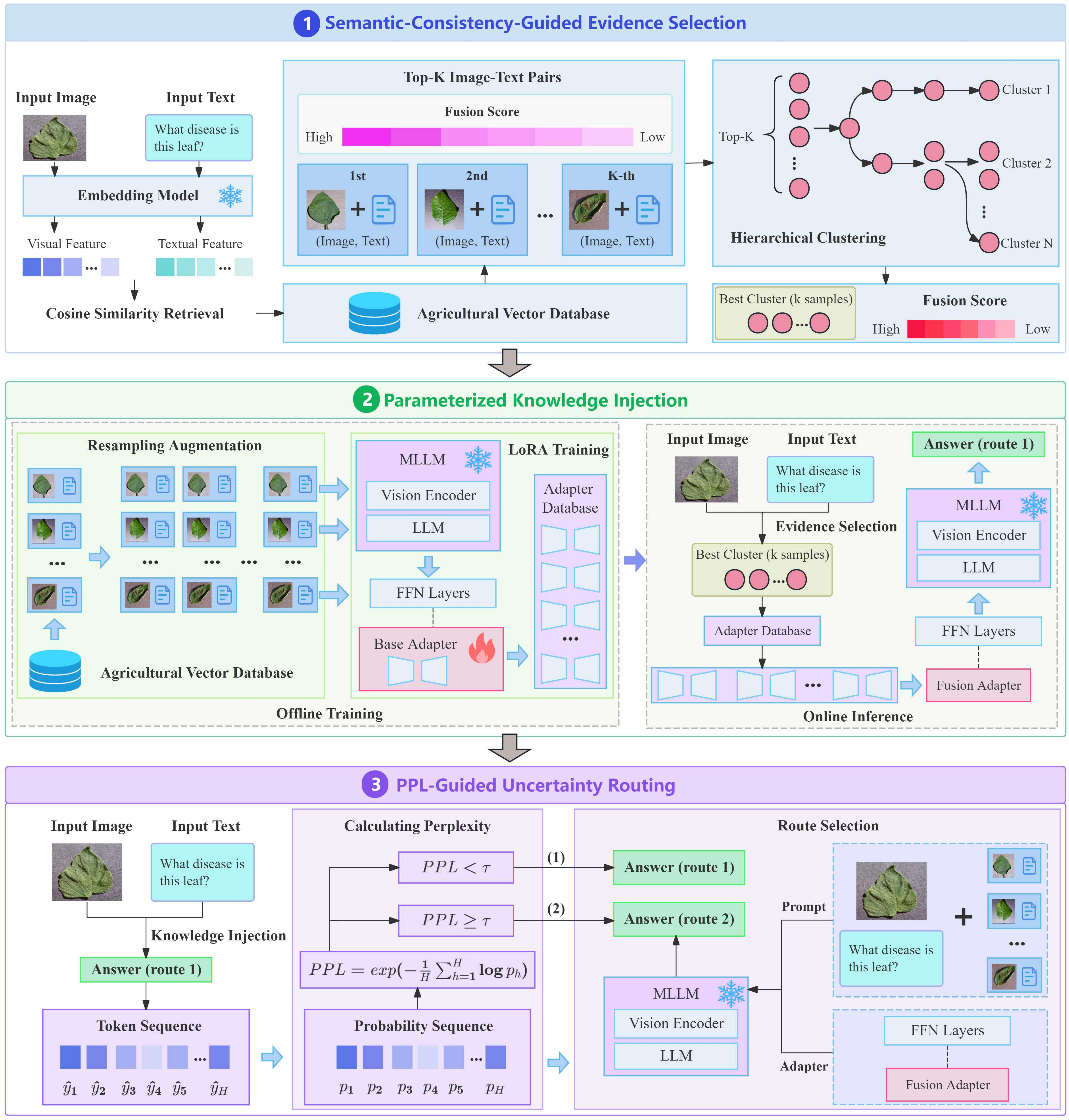}
\caption{Overview of the proposed Agri-MRAG framework. The framework first selects high-confidence evidence through semantic-consistency-guided clustering, then injects retrieved agricultural knowledge via parameterized adapters, and finally uses PPL-guided routing to switch between implicit parametric reasoning and explicit retrieval-assisted generation.}
\label{fig:agri_mrag}
\end{figure*}

The overall architecture of Agri-MRAG is illustrated in Figure~\ref{fig:agri_mrag}. {The overall workflow of the proposed Agri-MRAG method is presented in Algorithm~\ref{alg:agri_mrag}.} As an adaptive MRAG framework designed for the agricultural domain, Agri-MRAG consists of three core components: semantic-consistency-guided evidence selection, parameterized knowledge injection, and PPL-guided uncertainty routing, which are designed to mitigate retrieval noise and improve the utilization of domain knowledge. Specifically, given an agricultural image and a textual question, Agri-MRAG first retrieves an initial set of candidate image-text pairs from the agricultural knowledge base. The semantic-consistency-guided evidence selection mechanism then employs hierarchical clustering to identify high-confidence evidence that is semantically consistent with the query. Subsequently, the parameterized knowledge injection strategy dynamically assembles the corresponding parameterized adapters based on the selected evidence and injects agricultural knowledge into the MLLM to generate an answer. Finally, the PPL-guided uncertainty routing mechanism computes the PPL of the generated answer to estimate the model's generation uncertainty. When the uncertainty exceeds a predefined threshold, explicit retrieved evidence is incorporated as supplementary context. As a result, the model primarily relies on visual information and parametric knowledge for answer generation, while leveraging retrieved evidence for complementary reasoning in high-uncertainty scenarios, thereby improving the reliability of the generated responses.

\begin{algorithm}[t]

\caption{{Inference Procedure of Agri-MRAG}}
\label{alg:agri_mrag}
\begin{algorithmic}[1]
\Require Query image $I$, textual query $T$;
agricultural knowledge base $\mathcal{V}$;
LoRA adapter database $\mathcal{F}$;
candidate pool size $K$; number of clusters $N$;
modality weight $\alpha$; PPL threshold $\tau$
\Ensure Final answer $Y$

\vspace{0.5em}
\Statex \textbf{Stage I: Semantic-Consistency-Guided Evidence Selection}
\vspace{0.2em}

\State Extract query embeddings $v_I$ and $v_T$
\State Retrieve and rank candidate image--text pairs from $\mathcal{V}$
\State Select the top-$K$ candidates:
\[
\mathcal{D}=\{d_j=(I_j,T_j)\}_{j=1}^{K}
\]

\For{each pair $(d_m,d_n)\in\mathcal{D}\times\mathcal{D}$}
    \State Compute textual distance $D_{T}(d_m,d_n)$
    \State Compute visual distance $D_{I}(d_m,d_n)$
    \State Fuse the two distances:
    \[
    D(d_m,d_n)=\alpha D_{T}(d_m,d_n)
    +(1-\alpha)D_{I}(d_m,d_n)
    \]
\EndFor

\State Construct the distance matrix $\mathbf{D}_{mn} = [D(d_m, d_n)]$ 
\State Perform average-linkage agglomerative clustering
to obtain $\mathcal{C}=\{C_1,\ldots,C_N\}$
\For{each cluster $C_i\in\mathcal{C}$}
    \State Compute its mean relevance score $S(C_i)$
\EndFor
\State Select the evidence cluster:
\[
C^*=\arg\max_{C_i\in\mathcal{C}}S(C_i),
\qquad k=|C^*|
\]

\vspace{0.5em}
\Statex \textbf{Stage II: Parameterized Knowledge Injection}
\vspace{0.2em}

\State Retrieve the LoRA adapters associated with $C^*$ from $\mathcal{F}$
\State Fuse the selected adapters by factor concatenation
\State Apply the fused adapter update to the corresponding
feed-forward network parameters
\State Generate an initial answer $\hat{Y}$ using the adapted model
\State Compute the token-level negative log-likelihood
$\mathcal{L}_{gen}$ of $\hat{Y}$
\State Calculate the perplexity:
\[
\mathrm{PPL}=\exp(\mathcal{L}_{gen})
\]

\vspace{0.5em}
\Statex \textbf{Stage III: PPL-Guided Uncertainty Routing}
\vspace{0.2em}

\If{$\mathrm{PPL}<\tau$}
    \State $Y\gets\hat{Y}$
\Else
    \State Augment the query with the selected evidence $C^*$
    \State Generate the retrieval-assisted answer $Y$
    using the adapted model and explicit context
\EndIf

\State \Return $Y$
\end{algorithmic}
\end{algorithm}

\subsubsection{Semantic-Consistency-Guided Evidence Selection}
To reduce noise in retrieval results and select semantically consistent image-text pairs from the agricultural knowledge base, we propose a semantic-consistency-guided evidence selection mechanism. Specifically, given an input agricultural image $I$ and a textual query $T$, we first employ the embedding model to extract the corresponding visual and textual feature representations, denoted as $\mathbf{v}_I$ and $\mathbf{v}_T$, respectively. Subsequently, using $\mathbf{v}_I$ and $\mathbf{v}_T$ as retrieval queries, we perform cosine similarity-based search in the agricultural vector database $\mathcal{V}$. Moreover, we jointly consider the semantic similarity between the query text and retrieved text, as well as the visual similarity between the input image and retrieved images. For the $j$-th retrieved sample, the multimodal relevance score $s_j$ is defined in Equation \eqref{eq:cluster_1}:
\begin{equation}
s_j=\alpha \cdot \frac{\mathbf{v}_T \cdot \mathbf{v}_{T_j}}{\|\mathbf{v}_T\| \|\mathbf{v}_{T_j}\|} + (1-\alpha) \cdot \frac{\mathbf{v}_I \cdot \mathbf{v}_{I_j}}{\|\mathbf{v}_I\| \|\mathbf{v}_{I_j}\|}
\label{eq:cluster_1}
\end{equation}

where $\mathbf{v}_{T_j}$ and $\mathbf{v}_{I_j}$ denote the textual and visual feature vectors of the $j$-th retrieved image-text pair, respectively, and $\alpha$ is the textual modality weight. The Top-$K$ highest-scoring image-text pairs are then retained to form the candidate set: $\mathcal{D} = \{(I_j, T_j)\}_{j=1}^{K}$, where $I_j$ and $T_j$ denote the image and text of the $j$-th retrieved sample, respectively.

To further eliminate semantically inconsistent noisy retrievals, hierarchical clustering is applied to the candidate set $\mathcal{D}$. For any two samples $d_m = (I_m, T_m)$ and $d_n = (I_n, T_n)$, we first compute the cosine distances in both textual and visual modalities, as shown in Equation~\eqref{eq:cluster_2} and~\eqref{eq:cluster_3}:
\begin{equation}
D_T(d_m, d_n)=1 - \frac{\mathbf{v}_{T_m} \cdot \mathbf{v}_{T_n}} {\|\mathbf{v}_{T_m}\| \|\mathbf{v}_{T_n}\|}
\label{eq:cluster_2}
\end{equation}
\begin{equation}
D_I(d_m, d_n)=1 - \frac{\mathbf{v}_{I_m} \cdot \mathbf{v}_{I_n}}
{\|\mathbf{v}_{I_m}\| \|\mathbf{v}_{I_n}\|}
\label{eq:cluster_3}
\end{equation}

Considering the importance of visual information in Agri-VQA, a weighted fusion strategy is used to construct the pairwise distance as shown in Equation~\eqref{eq:cluster_4}, where $\alpha$ is the textual modality weight. {This yields a complete distance matrix: $\mathbf{D}_{mn} = [D(d_m, d_n)]$ and $m, n \in \{1, 2, \dots, K\}$, which represents the comprehensive multimodal distance between candidate image-text pairs $d_m$ and $d_n$, is symmetric, non-negative, and has a zero diagonal.}
\begin{equation}
D(d_m, d_n)=\alpha D_T(d_m, d_n)+(1-\alpha) D_I(d_m, d_n)
\label{eq:cluster_4}
\end{equation}

{Based on the distance matrix $\mathbf{D}_{mn}$, an agglomerative hierarchical clustering algorithm with average linkage is adopted in a bottom-up manner.} The inter-cluster distance between two clusters $C_p$ and $C_q$ is defined in Equation \eqref{eq:cluster_5}. Starting from singleton clusters, the algorithm iteratively merges the two closest clusters until a predefined number of clusters $N$ is obtained, forming the cluster set: $\mathcal{C} = \{C_1, C_2, \ldots, C_N\}$. For each cluster $C_i$ in $\mathcal{C}$, its relevance to the input query is defined as the average fusion relevance score of all samples within the cluster as shown in Equation~\eqref{eq:cluster_6}, where $|C_i|$ denotes the number of image-text pairs in $C_i$. The cluster with the highest score is selected as the final high-confidence knowledge subset: $C^{*} = \arg\max_{C_i} S(C_i)$. Unlike conventional fixed Top-$K$ retrieval strategies, the final number of retrieved contexts is adaptively determined by the size of the selected cluster: $k = |C^{*}|$.
\begin{equation}
D(C_p, C_q)=\frac{1}{|C_p||C_q|}\sum_{d_m \in C_p}\sum_{d_n \in C_q}D(d_m, d_n)
\label{eq:cluster_5}
\end{equation}
\begin{equation}
S(C_i)=\frac{1}{|C_i|}\sum_{d_j \in C_i}s_j
\label{eq:cluster_6}
\end{equation}

During the clustering process, we select only the single cluster that is most relevant to the query, since each cluster represents a distinct semantic distribution, and the best-matching cluster can be regarded as the semantic category to which the input query belongs. Based on this design, when the retrieved results exhibit high semantic consistency, more relevant knowledge within the selected cluster is retained; conversely, when the semantics are more dispersed, the context size is adaptively reduced, thereby effectively suppressing noisy information and improving robustness.

{Notably, our subsequent evaluation of the optimal number of clusters adopts downstream task performance, i.e., the BLEU, ROUGE-L, and METEOR scores of the model-generated answers, rather than internal clustering validation metrics such as the Silhouette coefficient.}
\subsubsection{Parameterized Knowledge Injection}

Through a semantic-consistency-guided evidence selection mechanism, a highly relevant subset of knowledge $C^{*}$ is retrieved from the agricultural knowledge base, and the final retrieval context size is adaptively determined as $k = |C^{*}|$. However, this knowledge still exists in the form of explicit image-text sequences. When directly fed into a MLLM via context concatenation, the external knowledge only participates in reasoning as short-term activated token sequences, and cannot be transformed into stable parameterized representations within the model, thereby limiting the model’s ability to deeply exploit domain-specific agricultural knowledge.

To further enhance knowledge utilization in the agricultural domain, Agri-MRAG introduces a parameterized knowledge injection strategy. This strategy encodes the agricultural knowledge contained in $C^{*}$ into LoRA adapters, enabling the knowledge to participate in the generation process in an implicit parametric form, thereby achieving more efficient multimodal fusion. The strategy consists of two stages: offline training and online inference.

In the offline training stage, each agricultural knowledge sample $d_j = (I_j, T_j)$ in the agricultural vector database $\mathcal{V}$ is treated as a minimal knowledge unit. Considering that supervision from a single sample is weak and may lead to unstable LoRA training, we perform $R$ rounds of resampling-based augmentation for each sample, constructing an augmented training set $D_j = \{d_j^{(1)}, d_j^{(2)}, \dots, d_j^{(R)}\}$. During training, the original parameters of the MLLM remain frozen, and only the low-rank matrices \( A_j \in \mathbb{R}^{r \times a} \) and \( B_j \in \mathbb{R}^{b \times r} \) corresponding to LoRA are optimized, where \( r \) denotes the matrix rank and satisfies \( r \ll \min(a,b) \). Correspondingly, the parameter increment for the FFN weight matrix \( \theta_{FFN} \in \mathbb{R}^{b \times a} \) is expressed as \( \Delta \theta_{j} = B_j A_j \).

The core motivation of this design lies in the strong instance dependency of agricultural knowledge. For example, the same disease may exhibit significant variations across different crops or growth stages. Therefore, a sample-level adapter enables finer-grained parameterization of agricultural knowledge. In addition, to ensure training stability and parameter-space consistency, all sample-specific adapters are fine-tuned from a shared base initialization. Ultimately, each knowledge sample corresponds to an independent parameterized adapter: $adapter_j = (A_j, B_j)$. These adapters form a parameterized adapter database $\mathcal{F} = \{adapter_j\}_{j=1}^{M}$, where $M$ denotes the total number of agricultural knowledge samples in $\mathcal{V}$.

In the online inference stage, the subset of adapters to be activated is first determined from the highly relevant knowledge subset $C^{*} = \{(I_j, T_j)\}_{j=1}^{k}$. This yields the adapter subset $F^{*} = \{adapter_j\}_{j=1}^{k}$, where each $adapter_j$ corresponds to the pair $(I_j, T_j)$ and $k = |C^{*}|$. Subsequently, parameter fusion is performed by concatenating operation on the low-rank matrices $(A_j, B_j)$ from all selected adapters, as shown in Equation~\eqref{eq:adapterr_1}. The spliced adapter is functionally equivalent to the linear superposition of multiple LoRA matrices, thereby enabling the dynamic combination and efficient injection of retrieved knowledge.

\begin{equation}
\Delta \theta_{fusion}=\begin{bmatrix}B_1 & B_2 & \cdots & B_k\end{bmatrix}
\begin{bmatrix}A_1 \\A_2 \\\vdots \\A_k\end{bmatrix}=\sum_{j=1}^{k} B_jA_j
\label{eq:adapterr_1}
\end{equation}

Finally, the fused incremental parameters are injected into the FFN layers of the MLLM, forming dynamically adapted model parameters conditioned on the current input, as shown in Equation~\eqref{eq:adapterr_2}. The MLLM then performs autoregressive generation based on the updated parameters, enabling deep integration of external agricultural knowledge at the parameter level.
\begin{equation}
\theta_{FFN}^{*} = \theta_{FFN} + \Delta \theta_{fusion}
\label{eq:adapterr_2}
\end{equation}

\subsubsection{PPL-Guided Uncertainty Routing} 
\label{analysis:ppl} 

The parameterized knowledge injection strategy enables the transformation of agricultural knowledge from explicit image--text sequences into internal parameterized representations of the MLLM, and achieves deep integration of domain knowledge through dynamic fusion adapters. However, different agricultural problems exhibit significant variation in their dependence on external knowledge. For tasks with clear semantics and relatively simple knowledge requirements, accurate reasoning can be achieved solely by relying on the parameterized adapter $F^{*}$. In contrast, for more complex agricultural scenarios such as fine-grained disease symptom discrimination, the implicit knowledge encoded in $F^{*}$ alone is often insufficient to capture all detailed information contained in the explicitly retrieved context $C^{*}$. On the other hand, although incorporating $C^{*}$ into all inference processes can provide richer evidence support, it also introduces additional computational overhead. Therefore, we introduce a PPL-guided uncertainty routing mechanism to dynamically assess the current cognitive state of the MLLM, determining whether $F^{*}$ is sufficient to solve the current problem and accordingly deciding whether to further incorporate $C^{*}$ for enhanced reasoning.

Specifically, the mechanism first applies the fusion adapter parameters $\Delta \theta_{fusion}$ to update the MLLM parameters $\theta$, resulting in the updated model parameters $\theta^{*}$. Then, using only the input agricultural image $I$ and textual query $T$, an implicit reasoning input is constructed as $X_{implicit} = (I, T)$. Without introducing the explicitly retrieved text $C^{*}$, the model performs first-stage inference based on $\theta^{*}$ to generate an initial response sequence $\hat{Y}$, as shown in Equation~\eqref{eq:ppl_1}, where $H$ denotes the length of the output sequence.
\begin{equation}
\hat{Y} = f(X_{implicit}; \theta^{*}) = \{\hat{y}_1, \hat{y}_2, \ldots, \hat{y}_H\}
\label{eq:ppl_1}
\end{equation}

For each generated token $\hat{y}_h$, the model outputs the corresponding conditional probability: $P(\hat{y}_h \mid X_{implicit}, \hat{y}_{<h}; \theta^{*})$. Based on the token probabilities along the generation trajectory, we measure the uncertainty of the current reasoning process using the average negative log-likelihood, as shown in Equation~\eqref{eq:ppl_2}. And the PPL is defined as Equation~\eqref{eq:ppl_3}:
\begin{equation}
\mathcal{L}_{gen} = -\frac{1}{H} \sum_{h=1}^{H} \log P(\hat{y}_h \mid X_{implicit}, \hat{y}_{<h}; \theta^{*})
\label{eq:ppl_2}
\end{equation}
\begin{equation}
PPL = \exp\left(\mathcal{L}_{gen}\right), \quad PPL \geq 1
\label{eq:ppl_3}
\end{equation}

The PPL is estimated from the actual generation path of the MLLM, reflecting the model's confidence in its prediction. A lower PPL indicates higher confidence, while a higher PPL suggests insufficient parameterized knowledge for reliable reasoning. Based on this, a threshold $\tau$ is introduced for PPL. When $PPL < \tau$, the model relies on parameterized knowledge and directly outputs $\hat{Y}$. When $PPL \geq \tau$, the model exhibits high uncertainty. In this case, the PPL-guided uncertainty routing mechanism constructs the explicit input $X_{explicit}$, as shown in Equation~\eqref{eq:ppl_4}. With this input, the model then performs inference, yielding $\hat{Y}^{*}$, as shown in Equation~\eqref{eq:ppl_5}.
\begin{equation}
X_{explicit} = X_{implicit} \cup C^{*} = (I, T) \cup \{(I_j, T_j)\}_{j=1}^{k}
\label{eq:ppl_4}
\end{equation}
\begin{equation}
\hat{Y}^{*} = f(X_{explicit}; \theta^{*}) = \{\hat{y}_1^{*}, \hat{y}_2^{*}, \ldots, \hat{y}_H^{*}\}
\label{eq:ppl_5}
\end{equation}

Ultimately, the PPL-guided uncertainty routing mechanism yields the final answer $Y$, where $Y \in \{\hat{Y}, \hat{Y}^{*}\}$. Through the joint effect of parameter-level domain priors and context-level fine-grained evidence, the model is able to perform complementary reasoning for complex agricultural problems, leading to more reliable final predictions.

\section{Results}
\subsection{Experimental Settings}

All experiments were implemented based on PyTorch 2.8.0 and conducted on a hardware platform equipped with 8 $\times$ NVIDIA GeForce RTX 4090 GPUs (24 GB memory per card) and 256 GB system RAM, with the Ubuntu 20.04.1 LTS operating system. We employ Qwen3-VL-2B~\cite{qwen3} as the base model and adopt a parameter-efficient fine-tuning strategy by injecting LoRA adapters into its FFN layers to acquire task-specific parametric knowledge. The text embedding model is BGE-M3~\cite{bge-m3}, and the image embedding model is BGE-VL-large~\cite{bge-vl}. The training procedure is configured with 1 epoch, a per-device batch size of 1 yielding a global batch size of 8, and the AdamW~\cite{adamw} optimizer with an initial learning rate of 0.0003. {The textual modality weight $\alpha$, an empirically determined parameter, is set to 0.3 and the number of clusters $N$ to 3, while Top-K is initialized at 10 as the candidate pool size, with the final retrieval count $k$ adaptively modulated by clustering.} The number of resampling rounds $R$ is set to 8, and the LoRA adapters are configured with a rank of 2 and a scaling factor of 32. The PPL threshold $\tau$ is set to 1.2. All experiments are conducted with frozen backbone weights, ensuring that only a minimal subset of parameters is trainable. The detailed experimental configuration is summarized in Table~\ref{env_config}.
\begin{table}[H]
	\caption{Experimental environment configuration.
    \centering
    \label{env_config}}
	\begin{tabularx}{\textwidth}{p{0.5\textwidth} p{0.5\textwidth}}
		\toprule
		\textbf{Experimental Environment} & \textbf{Experimental Configuration} \\
		\midrule
        Python Version & 3.10.19 \\
		Operating System & Ubuntu 20.04.1 LTS \\
		GPU & NVIDIA GeForce RTX 4090 $\times$ 8 \\
		GPU Memory & 24~GB per GPU \\
		System Memory & 256~GB \\
		Framework & PyTorch 2.8.0 \\
        Base MLLM & Qwen3-VL-2B \\
        Text Embedding Model & BGE-M3 \\
        Image Embedding Model & BGE-VL-large \\
        Number of Training Epochs & 1 \\
        Batch Size & 8 \\
        Optimizer & AdamW \\
        Initial Learning Rate & 0.0003 \\
        Maximum Retrieval Numbers Top-K & 10 \\
        The Textual Modality Weight $\alpha$ & 0.3 \\
        The Number of Clusters $N$ & 3 \\
        The Number of Resampling Rounds $R$ & 8 \\
        LoRA Rank & 2 \\
        LoRA Scaling Factor & 32 \\
        The Threshold $\tau$ of PPL & 1.2 \\
		\bottomrule
	\end{tabularx}
\end{table}

For dataset construction this study adopts PlantVillageVQA and PlantVQA; since PlantVillageVQA is quite large we randomly selected a subset of 158,977 question-answer pairs for experimental use. {Specifically, the full training split of PlantVillageVQA (154,977 question–answer pairs) was used, and 4,000 question–answer pairs were randomly sampled from its official test split using random seed 2026 as the test subset.} Table 2 shows that both datasets are split into training and test sets with each data sample being an image-text pair that comprises an image, a question, and an answer. The training set is used to construct the agricultural vector database and its corresponding parameterized adapter database whereas the test set serves to systematically evaluate Agri‑MRAG and its comparative methods so as to validate the effectiveness of the proposed approach on Agri‑VQA tasks.

\begin{table}[H]
  \caption{The partition details of the PlantVillageVQA and PlantVQA datasets.
  \centering 
  \label{tab:dataset}}
  \small
  \renewcommand{\arraystretch}{1.0} 
  \begin{tabularx}{\linewidth}{CCCC}
    \toprule
    \textbf{Dataset} & \textbf{Train Set} & \textbf{Test Set} & \textbf{Total} \\
    \midrule
    PlantVillageVQA & 154,977 & 4,000 & 158,977 \\
    \# Percentage & 97.48\% & 2.52\% & - \\
    \midrule
    PlantVQA & 12,038 & 2,990 & 15,028 \\
    \# Percentage & 80.10\% & 19.90\% & - \\
    \bottomrule
  \end{tabularx}
\end{table}

\subsection{Evaluation Metrics}
ri‑VQA task, this section quantitatively examines model outputs in terms of semantic consistency. Hence we employ BLEU~\cite{bleu}, ROUGE‑L~\cite{rouge}, and METEOR~\cite{meteor} as our evaluation metrics, which measure how well generated responses align with ground‑truth answers grounded in agricultural knowledge. Among these, BLEU operates on $n$-gram precision matching with a brevity penalty, primarily focusing on local lexical overlap accuracy. In this work we adopt the average BLEU score, computed as the arithmetic mean of BLEU‑1 through BLEU‑4, to achieve a more balanced assessment across different $n$-gram orders. ROUGE‑L relies on the longest common subsequence (LCS) and balances precision and recall to capture both global structural consistency and sequential coherence in lengthy agricultural descriptions. METEOR meanwhile incorporates stemming and synonym matching and also applies a fragmentation penalty, thereby effectively reducing semantic deviations that arise from pure string‑based matching when agricultural terminology varies.

{It should be noted that BLEU, ROUGE-L, and METEOR mainly measure textual consistency with reference answers rather than diagnostic correctness itself. A high score indicates that the generated response is lexically or semantically close to the reference wording. These metrics should therefore be regarded as complementary indicators of response quality, and a more comprehensive assessment of diagnostic correctness would require expert evaluation, which we leave as an important direction for future work.}

\subsection{Comparisons with Other Methods}

To evaluate Agri-MRAG, we compare it with three categories of baseline methods. For general-purpose MLLMs, we include Qwen3-VL-4B~\cite{qwen3}, Qwen3-VL-8B~\cite{qwen3}, LLaVA-OneVision-2-8B~\cite{llava}, and GLM-4.1V-9B-Thinking~\cite{glm4.6v}. These models represent open-source MLLMs with different model scales and reasoning capabilities. For agriculture-specific MLLMs, we use AgriChat~\cite{agrichat}, which is fine-tuned on the AgriMM dataset for agricultural tasks such as disease diagnosis and crop identification. To assess the effect of retrieval augmentation, we also include mRAG~\cite{mrag} and VisRAG~\cite{visrag}, whose underlying MLLM and external knowledge base are kept strictly consistent with those of Agri-MRAG to ensure a fair comparison.

The detailed performance comparison results of each model on the two datasets are presented in Table~\ref{sota}. Based on a systematic analysis of the experimental results, the following conclusions can be drawn.

\begin{table}[H]
    \caption{Performance comparison between our method and other approaches. Best results are in \textbf{bold}.\label{sota}}
    \footnotesize
    \renewcommand{\arraystretch}{1.35}
    
    \resizebox{\textwidth}{!}{%
    \begin{tabular}{l c c ccc ccc}
        \toprule
        \multirow{2}{*}{\makecell{Model}} 
        & \multirow{2}{*}{\makecell{{Params(B)}}} 
        & \multirow{2}{*}{\makecell{{GFLOPS}}} 
        & \multicolumn{3}{c}{PlantVillageVQA} 
        & \multicolumn{3}{c}{PlantVQA} \\
        \cmidrule(lr){4-6} \cmidrule(l){7-9}
        & & & BLEU & ROUGE-L & METEOR & BLEU & ROUGE-L & METEOR \\
        \midrule
        Qwen3-VL-4B~\cite{qwen3}            & {4} & {1{,}200} & 12.74 & 34.34 & 19.12 & 21.57 & 38.09 & 28.53 \\
        Qwen3-VL-8B~\cite{qwen3}            & {8} & {2{,}400} & 13.79 & 34.53 & 19.56 & 22.90 & 43.41 & 29.43 \\
        LLaVA-OneVision-2-8B~\cite{llava}   & {8} & {2{,}500} & 4.94  & 14.02 & 14.41 & 4.60  & 1.50  & 17.49 \\
        GLM-4.1V-9B-Thinking~\cite{glm4.6v} & {9} & {2{,}700} & 13.42 & 33.08 & 19.25 & 18.57 & 34.01 & 27.96 \\
        AgriChat (7B)~\cite{agrichat}       & {7} & {2{,}100} & 7.60  & 21.57 & 23.58 & 5.51  & 3.94  & 17.50 \\
        mRAG~\cite{mrag}                    & {2} & {600}     & 11.91 & 29.51 & 19.49 & 21.30 & 42.07 & 27.98 \\
        VisRAG~\cite{visrag}                & {2} & {600}     & 13.35 & 32.41 & 20.91 & 22.51 & 43.19 & 29.60 \\
        Agri-MRAG (\textbf{Ours})           & {2} & {600}     & \textbf{16.71} & \textbf{41.21} & \textbf{25.62} & \textbf{24.29} & \textbf{46.58} & \textbf{31.73} \\
        \bottomrule
    \end{tabular}%
    }
\end{table}
\begin{table}[H]
    
    \centering
    \small
    \caption{{Comparison of comparator controls and implementation settings among mRAG, VisRAG, and Agri-MRAG. All methods use the same base MLLM (Qwen3-VL-2B) and the same external agricultural knowledge base to ensure a fair comparison. The specific details of the update strategy can be found in lines 816–818.}}
    \label{comparator_controls}
    \renewcommand{\arraystretch}{1.3}

    \resizebox{\textwidth}{!}{%
    \begin{tabular}{lccc}
        \toprule
        \textbf{Item} & \textbf{mRAG} & \textbf{VisRAG} & \textbf{Agri-MRAG} \\
        \midrule
        Retrieval Budget (Top-K) & 10 & 10 & 10 (adaptive $k$) \\
        Prompt Length (tokens/sample) & 197.41 & 197.15 & 179.10 \\
        Context Length (tokens/sample) & 172.41 & 172.15 & 71.71 \\
        Training / Tuning Procedure & No training & No training & LoRA adapter training \\
        Decoding Settings & Greedy decoding & Greedy decoding & Greedy decoding \\
        Compute Budget & Retrieval + Inference & Retrieval + Inference & Adapter Training + Retrieval + Inference \\
        \bottomrule
    \end{tabular}%
    }
\end{table}

Despite possessing 8 billion and 9 billion parameters respectively, Qwen3-VL-8B and GLM-4.1V-9B-Thinking do not significantly outperform the smaller parameter Qwen3-VL-4B on fine-grained agricultural pest and disease diagnosis tasks, and even exhibit declines on several metrics. This indicates that simply increasing parameter counts or relying on general pre-training knowledge cannot compensate for the inherent lack of domain-specific expertise in general-purpose models. In contrast, Agri-MRAG substantially surpasses all these general baselines across all three evaluation metrics. Specifically, on the PlantVillageVQA dataset, Agri-MRAG achieves a ROUGE-L score of 41.21, which is 6.68 points higher than that of the strongest general baseline, Qwen3-VL-8B, at 34.53. This demonstrates that introducing an external agricultural knowledge base and performing deep fusion via a parameterized adapter can effectively mitigate the hallucination problem of general MLLMs in understanding specialized terminology and describing pathological features.

Compared to the domain-specific baseline AgriChat, Agri-MRAG also exhibits significant advantages. AgriChat, as a state-of-the-art model fine-tuned on large-scale agricultural multi-modal datasets, performs moderately on the METEOR metric but markedly underperforms Agri-MRAG on BLEU and ROUGE-L. For example, on the PlantVQA dataset, Agri-MRAG achieves a BLEU score of 24.29, which is more than four times that of AgriChat’s corresponding score of 5.51, and its ROUGE-L score of 46.58 far exceeds AgriChat’s 3.94. Such a substantial performance gap suggests that while traditional supervised fine-tuning allows a model to memorize specific agricultural patterns, the lack of dynamic retrieval and knowledge augmentation mechanisms severely limits its generalization capability when faced with unseen query scenarios. Agri-MRAG, through its dynamic routing and cluster selection mechanism, retrieves the most relevant evidence in real time according to specific questions, thereby achieving considerable performance improvements.

Compared with existing retrieval-augmented baseline methods mRAG and VisRAG, Agri-MRAG still maintains a performance advantage. Experimental results show that on the PlantVillageVQA dataset, Agri-MRAG achieves BLEU, ROUGE-L, and METEOR scores of 16.71, 41.21, and 25.62, respectively, comprehensively surpassing the corresponding scores of mRAG, which are 11.91, 29.51, and 19.49, and VisRAG, which are 13.35, 32.41, and 20.91. On the PlantVQA dataset, Agri-MRAG attains BLEU, ROUGE-L, and METEOR scores of 24.29, 46.58, and 31.73, also outperforming mRAG with scores of 21.30, 42.07, and 27.98, and VisRAG with scores of 22.51, 43.19, and 29.60. Notably, on the PlantVillageVQA dataset, Agri-MRAG leads mRAG and VisRAG by 11.70 and 8.80 points, respectively, in terms of the ROUGE-L metric. This substantial gap indicates that simple generic retrieval-augmented strategies struggle to effectively address the complex fine-grained visual–language alignment problem in agricultural scenarios. In contrast, through a domain-specific cluster selection mechanism and adapter injection, Agri-MRAG can more precisely locate evidence highly relevant to agricultural pathological features, thereby generating more accurate diagnostic answers.

{Furthermore, Table~\ref{comparator_controls} summarizes the key implementation settings and comparator controls across mRAG, VisRAG, and Agri-MRAG. While all methods share the same base MLLM (Qwen3-VL-2B) and external agricultural knowledge base with a Top-K retrieval budget of 10, their processing pipelines differ. Unlike mRAG and VisRAG, which rely solely on prompt-level concatenation without additional training, Agri-MRAG incorporates an offline LoRA adapter training phase. Notably, despite this extra training cost and a comparable prompt length, Agri-MRAG maintains a significantly shorter context length (71.71 tokens/sample) compared to the baselines (over 172 tokens/sample). This reduction is attributed to the PPL-guided routing mechanism, which avoids unnecessary explicit context expansion for confident queries. Based on these aligned settings, the following sections evaluate the performance benefits of our framework.}

Notably, LLaVA-OneVision-2-8B and AgriChat perform extremely poorly on the PlantVQA dataset, with ROUGE-L scores of only 1.50 and 3.94 respectively, yet achieve scores of 14.02 and 21.57 respectively on the PlantVillageVQA dataset. This stark performance gap results from the two models' failure to adequately align agricultural fine-grained features across different datasets. In contrast, Agri-MRAG maintains leading and stable high performance on both datasets with different distributions, demonstrating its excellent domain adaptation capability.

\subsection{Ablation Experiments}
\subsubsection{Ablation on Key modules}

To systematically evaluate the contributions of each key module, we conducted ablation studies. We gradually introduced each module and observed the resulting performance changes, thereby quantitatively evaluating how every module affected the visual question answering outcomes, with the detailed results presented in Table~\ref{ablation}.

We compare the baseline model against the one that only incorporates the Cluster module and observe a clear performance improvement. On the PlantVillageVQA dataset, BLEU rises from {11.75} to {14.20}, ROUGE‑L from {29.10} to {34.93}, and METEOR from {19.25} to {22.31}. This notable gain indicates that by carrying out semantically consistent clustering and selecting high‑confidence subsets, Agri‑MRAG effectively filters out irrelevant noise and hence ensures a strong relevance between the input knowledge pieces and the query.
\begin{table}[H]
    
    \centering
    \small
    \caption{Ablation study of adding Cluster, Adapter and PPL. The first row corresponds to the baseline Qwen3-VL-2B. The last row refers to the complete design of the Agri-MRAG. {Our results are reported as mean $\pm$ standard deviation over 3-fold cross-validation.} Best results are in \textbf{bold}.\label{ablation}}
    \renewcommand{\arraystretch}{1.2}

    \resizebox{\textwidth}{!}{%
    \begin{tabular}{ccc ccc ccc}
        \toprule
        \multicolumn{3}{c}{Module} & \multicolumn{3}{c}{PlantVillageVQA} & \multicolumn{3}{c}{PlantVQA} \\
        \cmidrule(lr){1-3} \cmidrule(lr){4-6} \cmidrule(l){7-9}
        Cluster & Adapter & PPL & BLEU & ROUGE-L & METEOR & BLEU & ROUGE-L & METEOR \\
        \midrule
        & & 
        & 11.75 & 29.10 & 19.25 & 21.14 & 41.71 & 27.91 \\
        & & 
        & ($\pm$0.11) & ($\pm$0.30) & ($\pm$0.20) & ($\pm$0.13) & ($\pm$0.14) & ($\pm$0.14) \\

        \multirow{2}{*}{\checkmark} & & 
        & 14.20 & 34.93 & 22.31 & 23.33 & 45.59 & 30.57 \\
        & & 
        & ($\pm$0.23) & ($\pm$0.26) & ($\pm$0.34) & ($\pm$0.29) & ($\pm$0.34) & ($\pm$0.15) \\

        \multirow{2}{*}{\checkmark} & \multirow{2}{*}{\checkmark} & 
        & 16.05 & 38.88 & 24.42 & 23.82 & 46.13 & 31.11 \\
        & & 
        & ($\pm$0.37) & ($\pm$0.32) & ($\pm$0.24) & ($\pm$0.21) & ($\pm$0.28) & ($\pm$0.24) \\

        \multirow{2}{*}{\checkmark} & \multirow{2}{*}{\checkmark} & \multirow{2}{*}{\checkmark}
        & \textbf{16.54} & \textbf{40.75} & \textbf{25.34} & \textbf{24.02} & \textbf{46.26} & \textbf{31.50} \\
        & & 
        & ($\pm$0.05) & ($\pm$0.06) & ($\pm$0.08) & ($\pm$0.12) & ($\pm$0.27) & ($\pm$0.23) \\
        \bottomrule
    \end{tabular}%
    }
\end{table}

{Introducing the Adapter module on top of the Cluster module yields a substantial performance leap. Take PlantVillageVQA as an example: BLEU rises from 14.20 to 16.05, ROUGE-L from 34.93 to 38.88, and METEOR from 22.31 to 24.42. This remarkable gain reveals the true significance of reshaping the knowledge fusion paradigm. Traditional text concatenation often expands the context window and diverts attention; by contrast, the Adapter approach turns explicit agricultural knowledge into parametric bias embedded in the model. This conversion enables the model to call upon deep agronomic expertise at a lower computational expense during subsequent reasoning steps.}

{Once the PPL dynamic routing mechanism is fully integrated, Agri-MRAG achieves the best performance across all evaluation metrics on both datasets. On the PlantVillageVQA dataset, the ROUGE-L score further climbs from 38.88 to 40.75, and the METEOR score increases from 24.42 to 25.34. On the PlantVQA dataset, it also attains the highest BLEU (24.02), ROUGE-L (46.26), and METEOR (31.50) scores.} This gain reveals the core strength of the PPL mechanism, namely the on‑demand allocation of computational resources. When facing simple questions or cases where the model already holds sufficient confidence, the PPL mechanism lets the model skip the costly explicit text verification steps and instead directly draw on parametric knowledge for rapid answering, thus preserving efficiency. For highly uncertain case diagnoses, PPL triggers a collaborative mode, using fine-grained retrieval evidence to correct potential computational biases.

The ablation study reveals the complementary relationship among the three modules; the absence of any module imposes varying degrees of constraint on system performance. Without the Cluster module, the model lacks sufficient agricultural knowledge support. Without the Adapter module, the model cannot fully utilize external knowledge. Without the PPL module, the model may over-rely on lengthy explicit context for simple questions or lack sufficient evidentiary support for complex questions.

\subsubsection{Ablation on Top-K}

To analyze the impact of the maximum retrieval numbers Top-K on the performance of Agri-MRAG, we conducted ablation experiments on the PlantVillageVQA and PlantVQA datasets, with $K=\{0,5,10,15,20\}$. Table~\ref{top-k} shows the experimental results; overall, model performance first improves markedly with increasing Top‑K and then levels off. With $k=0$, the model relies entirely on its parametric knowledge without external retrieval, which yields the lowest scores on both datasets. When more items are retrieved, the model obtains richer and more relevant agricultural knowledge, thus bringing notable gains in BLEU, ROUGE‑L, and METEOR.

\begin{table}[H]
    \caption{Ablation study on the effect of different maximum retrieval numbers Top-K. Best results are shown in \textbf{bold}. {ITPS denotes the average inference time per sample (in seconds), and Token\_doc denotes the average total token length of the documents retrieved for each data instance.}\label{top-k}}
    \small
    \renewcommand{\arraystretch}{1.5} 
    
    \resizebox{\textwidth}{!}{%
    \begin{tabular}{c ccc cc ccc cc}
        \toprule
        \multirow{2}{*}{\makecell{Top-K}} & \multicolumn{5}{c}{PlantVillageVQA} & \multicolumn{5}{c}{PlantVQA} \\
        \cmidrule(lr){2-6} \cmidrule(l){7-11}
        & BLEU & ROUGE-L & METEOR & {ITPS} & {Token\_doc} & BLEU & ROUGE-L & METEOR & {ITPS} & {Token\_doc} \\
        \midrule
        $K=0$   & 11.87 & 29.45 & 19.48 & {\textbf{9.78}} & {\textbf{0.00}} & 21.00 & 41.60 & 27.77 & {\textbf{1.85}} & {\textbf{0.00}} \\
        $K=5$   & 15.91 & 37.96 & 23.86 & {12.39} & {30.10} & 23.34 & 45.75 & 30.63 & {5.90} & {28.56} \\
        $K=10$  & 16.71 & 41.21 & 25.62 & {13.92} & {70.07} & \textbf{24.29} & \textbf{46.58} & \textbf{31.73} & {7.69} & {73.91} \\
        $K=15$  & 18.68 & 42.34 & 26.63 & {15.37} & {116.53} & 23.88 & 46.35 & 31.16 & {9.33} & {124.05} \\
        $K=20$  & \textbf{19.13} & \textbf{43.04} & \textbf{27.20} & {16.90} & {165.03} & 24.08 & \textbf{46.58} & 31.23 & {10.88} & {178.38} \\
        \bottomrule
    \end{tabular}%
    }
\end{table}

{The results also indicate that different datasets respond differently to varying Top-K values. On the PlantVillageVQA dataset, with $K=0$, the BLEU, ROUGE-L, and METEOR scores stand at 11.87, 29.45, and 19.48, respectively; when $K$ reaches 20, these three metrics rise to 19.13, 43.04, and 27.20, corresponding to improvements of 7.26, 13.59, and 7.72.} This continuous upward trend across all metrics as Top-K grows suggests that agricultural knowledge within this dataset is highly complementary, hence retrieving more items consistently supplies effective information to the model and the best results are achieved at $K=20$. For the PlantVQA dataset, however, the situation differs: with $K=0$, BLEU, ROUGE-L, and METEOR are 21.00, 41.60, and 27.77, respectively; at $K=10$, these scores climb to 24.29, 46.58, and 31.73, representing gains of 3.29, 4.98, and 3.96. Model performance essentially peaks at $K=10$, where BLEU and METEOR reach their highest values and ROUGE-L ties its best result at both $K=10$ and $K=20$. This pattern implies that once the retrieval number passes a certain threshold, further enlarging the retrieved set yields only marginal gains and may even introduce redundant information that slightly hurts some metrics. Thus, an appropriate control over the retrieval scale not only ensures sufficient knowledge coverage but also prevents irrelevant knowledge from interfering with the model's reasoning.

{Beyond these performance results, we also examine inference efficiency, which is measured by the average end-to-end inference latency per sample on the test set, expressed in seconds.} {Raising Top-K from 10 to 20 brings in extra context. This is clearly reflected in the token length of all retrieved documents per data instance (denoted as Token\_doc), which increases substantially from 70.07 to 165.03 on PlantVillageVQA and from 73.91 to 178.38 on PlantVQA. Consequently, this expanded context leads to an average per-sample inference time (ITPS) increase of 2.98 s on PlantVillageVQA and 3.18 s on PlantVQA.} {Unless otherwise stated, all inference times reported in this paper are measured end-to-end.} With a large test set, this additional overhead accumulates substantially and leads to considerable latency, while the corresponding performance improvements have become very modest. Hence we opt for $K=10$ as the optimal setting after balancing model performance against efficiency considerations.

\subsubsection{Ablation on Cluster Number}
To investigate how the cluster number affects Agri-MRAG's performance, we carry out ablation experiments on both the PlantVillageVQA and PlantVQA datasets by varying $N$, and the outcomes are reported in Table~\ref{cluster}.

Looking at the data trends, we find that when the cluster number grows from $N=1$ to $N=3$, the evaluation metrics on both datasets generally increase and reach their peaks at $N=3$. On PlantVillageVQA, the scores at this setting are 16.71, 41.21, and 25.62, which correspond to gains of 0.32, 0.92, and 0.21 over $N=1$. On PlantVQA, the best results also occur at $N=3$, giving 24.29, 46.58, and 31.73, with improvements of 0.40, 0.30, and 0.56 respectively. These observations suggest that choosing a suitable cluster number helps retrieve the most relevant context and thus strengthens the model's ability to aggregate contextual information and align multimodal semantics.

\begin{table}[H]
    \caption{Ablation study on the effect of different cluster numbers $N$. Best results are shown in \textbf{bold}.\label{cluster}}
    \small
    \renewcommand{\arraystretch}{1.5} 
    
    \begin{tabularx}{\textwidth}{>{\centering\arraybackslash}X ccc ccc}
        \toprule
        \multirow{2}{*}{\makecell{Cluster Number}} & \multicolumn{3}{c}{PlantVillageVQA} & \multicolumn{3}{c}{PlantVQA} \\
        \cmidrule(lr){2-4} \cmidrule(l){5-7}
        & BLEU & ROUGE-L & METEOR & BLEU & ROUGE-L & METEOR \\
        \midrule
        $N=1$   & 16.39 & 40.29 & 25.41 & 23.89 & 46.28 & 31.17 \\
        $N=2$   & 16.52 & 40.55 & 25.58 & 23.70 & 46.18 & 30.94 \\
        $N=3$   & \textbf{16.71} & \textbf{41.21} & \textbf{25.62} & \textbf{24.29} & \textbf{46.58} & \textbf{31.73} \\
        $N=4$   & 16.49 & 40.53 & 25.51 & 23.61 & 45.75 & 30.87 \\
        \bottomrule
    \end{tabularx}
\end{table}

Further increasing the cluster number to $N=4$ nonetheless results in a notable performance decline. Compared to $N=3$, the BLEU, ROUGE-L, and METEOR scores on the PlantVillageVQA dataset decreased by 0.22, 0.68, and 0.11, respectively. Similarly, the three metrics on the PlantVQA dataset experienced more substantial decreases of 0.68, 0.83, and 0.86, respectively. This phenomenon suggests that excessive clustering leads to overly fragmented contexts, reducing the amount of valid information and consequently impairing the model's overall performance. Based on the above analysis, the optimal cluster number for our study is set to $N=3$.

\subsubsection{Ablation on PPL Threshold}
PPL threshold $\tau$ on the Agri-MRAG method, we conducted ablation experiments on the PlantVillageVQA and PlantVQA datasets, with the experimental results shown in Table~\ref{Threshold}. The threshold $\tau$ controls the routing of inference. When the perplexity of a sample is below the threshold, only the parametric adapter is used for inference; otherwise, both the parametric adapter and explicit context are employed jointly. In particular, $\tau=1.0$ indicates that all samples in the test set use the joint inference of the parametric adapter and explicit context.

\begin{table}[H]
    \caption{Ablation study on the effect of different PPL threshold $\tau$. Best results are shown in \textbf{bold}. {ITPS denotes the average inference time per sample (in seconds).}\label{Threshold}}
    \scriptsize
    \renewcommand{\arraystretch}{1.5} 
    
    \begin{tabularx}{\textwidth}{>{\centering\arraybackslash}X ccc c ccc c}
        \toprule
        \multirow{2}{*}{\makecell{PPL Threshold}} & \multicolumn{4}{c}{PlantVillageVQA} & \multicolumn{4}{c}{PlantVQA} \\
        \cmidrule(lr){2-5} \cmidrule(l){6-9}
        & BLEU & ROUGE-L & METEOR & {ITPS (s)} & BLEU & ROUGE-L & METEOR & {ITPS (s)} \\
        \midrule
        $\tau=1.0$   & \textbf{19.69} & \textbf{45.76} & \textbf{28.18} & {15.49} & \textbf{24.51} & \textbf{46.63} & \textbf{32.11} & {10.89} \\
        $\tau=1.2$   & 16.71 & 41.21 & 25.62 & {15.03} & 24.29 & 46.58 & 31.73 & {10.41} \\
        $\tau=1.4$   & 14.43 & 34.80 & 22.25 & {\textbf{14.61}} & 23.35 & 45.61 & 30.59 & {10.27} \\
        $\tau=1.6$   & 13.28 & 32.55 & 20.96 & {15.05} & 23.32 & 45.65 & 30.53 & {\textbf{10.21}} \\
        \bottomrule
    \end{tabularx}
\end{table}

\begin{table}[H]
    
    \caption{{Routing proportions and latency distributions under different PPL thresholds. Route 1 denotes PPL $\leq \tau$, and Route 2 denotes PPL $> \tau$. ITPS1 and ITPS2 are the average inference time per sample (in seconds) in Route 1 and Route 2, respectively.}\label{route}}
    \scriptsize
    \renewcommand{\arraystretch}{1.5} 
    
    \begin{tabularx}{\textwidth}{>{\centering\arraybackslash}X ccc c ccc c}
        \toprule
        \multirow{2}{*}{\makecell{PPL Threshold}} & \multicolumn{4}{c}{PlantVillageVQA} & \multicolumn{4}{c}{PlantVQA} \\
        \cmidrule(lr){2-5} \cmidrule(l){6-9}
        & Route1 & Route2 & ITPS1 (s) & {ITPS2 (s)} & Route1 & Route2 & ITPS1 (s) & {ITPS2 (s)} \\
        \midrule
        $\tau=1.0$   & 1166 & 2834 & 15.41 & 15.66 & 1671 & 1319 & 10.85 & 10.95 \\
        $\tau=1.2$   & 1535 & 2465 & 14.72 & 15.22 & 2154 & 836 & 10.32 & 10.62 \\
        $\tau=1.4$   & 3095 & 905 & 14.38 & 15.40& 2569 & 421 & 10.26 & 10.29 \\
        $\tau=1.6$   & 3837 & 163 & 14.97 & 16.98 & 2792 & 198 & 10.19 & 10.46 \\
        \bottomrule
    \end{tabularx}
\end{table}

From the experimental results, as $\tau$ increases from 1.0 to 1.6, the overall performance on both datasets exhibits a downward trend, but the magnitude of decline differs substantially. On the PlantVillageVQA dataset, the model is more sensitive to threshold variations. When $\tau$ increases from 1.0 to 1.2, BLEU, ROUGE-L, and METEOR drop by 2.98, 4.55, and 2.56, respectively; when $\tau$ further increases to 1.6, the three metrics drop cumulatively by 6.41, 13.21, and 7.22. This implies that as the threshold increases, an increasing number of samples stop engaging explicit context and instead rely only on the parametric adapter for inference; consequently, the model struggles to fully exploit domain knowledge stored in the agricultural knowledge base, which leads to a steady drop in generation quality. On the PlantVQA dataset, however, performance changes are much more moderate. When $\tau$ rises from 1.0 to 1.2, BLEU, ROUGE-L, and METEOR decline by just 0.22, 0.05, and 0.38, respectively; even when $\tau$ reaches 1.6, the decreases relative to $\tau=1.0$ are only 1.19, 0.98, and 1.58. This observation implies that only a small subset of PlantVQA samples actually benefits from explicit context, whereas the majority can be adequately handled by the parametric adapter alone, hence the threshold variation exerts a fairly limited impact on overall performance.

Considering both performance and inference efficiency, although $\tau=1.0$ achieves the best performance on both datasets, it requires loading both the parametric adapter and explicit context for every test sample, leading to a notable increase in inference time. {Specifically, on the PlantVillageVQA dataset, the average inference time at $\tau = 1.0$ is 0.46~s longer than that at $\tau = 1.2$; on the PlantVQA dataset, the average inference time increases by 0.48~s. In comparison, $\tau = 1.2$ preserves almost the same performance as $\tau = 1.0$ on PlantVQA, and although a certain performance drop exists on PlantVillageVQA, it effectively reduces inference cost. This reflects a clear quality--efficiency trade-off: lowering the threshold to $\tau = 1.0$ improves BLEU, ROUGE-L, and METEOR at the cost of longer inference time, whereas $\tau = 1.2$ achieves a more favorable balance by maintaining comparable performance on PlantVQA and acceptable performance on PlantVillageVQA with lower inference cost. Therefore, we ultimately select $\tau = 1.2$ as an appropriate PPL threshold.}

{In addition, Table \ref{route} further shows the routing proportions and latency on the two datasets. At \(\tau = 1.2\), the numbers of samples routed to Route~1 and Route~2 are 1535 and 2465 on PlantVillageVQA, respectively, and 2154 and 836 on PlantVQA, respectively. In terms of inference time, at \(\tau = 1.2\), the average per-sample inference time of Route~1 and Route~2 is 14.72~s and 15.22~s on PlantVillageVQA, respectively, and 10.32~s and 10.62~s on PlantVQA, respectively. The average inference time of Route~2 is higher than that of Route~1 on both datasets.}

{It should be noted that no independent validation partition was used in this work. The values of $K$, $N$, and $\tau$ were determined on the test sets of the current datasets by comparing the framework's behavior under different thresholds, with the aim of illustrating its behavior across thresholds and providing a reference for determining a suitable threshold range for these datasets. When applying the framework to other datasets, $K$, $N$, and $\tau$ should be selected based on an independent validation set rather than on the final test results, which we leave as a standard practice for future work.}

\subsection{Case Study}
In the evaluation of VQA tasks within the agricultural domain, this study selected two representative datasets, PlantVillageVQA and PlantVQA, as test benchmarks, as illustrated in Figure~\ref{case}.

Case (a) presents a leaf disease identification task involving squash from the PlantVillageVQA dataset, where the ground-truth answer is "Powdery mildew". In comparative experiments, general-purpose MLLMs exhibited significant limitations. Specifically, both the 4B and 8B parameter versions of the Qwen3-VL series incorrectly identified the symptoms as "Downy mildew", while LLaVA-OneVision-2-8B misclassified them as "Bacterial leaf spot". Despite being fine-tuned on agricultural data, AgriChat nonetheless failed to overcome the cognitive biases inherent in general models and accordingly predicted ``Downy mildew'' as well. By contrast, those methods that employ retrieval-augmented mechanisms—namely mRAG, VisRAG, and Agri-MRAG—along with the GLM-4.1V-9B-Thinking model that possesses deep reasoning capabilities all correctly output ``Powdery mildew''. This outcome strongly suggests that without external knowledge base support even large-parameter general models tend to confuse diseases that are morphologically similar, whereas retrieval augmentation can offset the deficiencies of internal parametric knowledge and thereby enhance diagnostic accuracy.

Case (b) turns to a fruit surface symptom recognition task from the PlantVQA dataset, where the goal is to identify red patch-like abnormalities on peach surfaces and the ground-truth label is ``Peach scab''. Here the errors made by different models exhibit diverse patterns. Qwen3-VL-4B reaches a diagnosis of ``Peach brown spot disease'' while Qwen3-VL-8B gives ``Brown rot of peach'', which reflects instability when the models process fruit epidermal texture features. LLaVA-OneVision-2-8B does point to symptoms consistent with ``Peach scab'' yet leaves its description rather vague. The GLM-4.1V-9B-Thinking model, on the other hand, suffers from hallucination and wrongly associates the symptoms with ``Bacterial shot hole disease'', suggesting that chain-of-thought reasoning alone—without accurate visual grounding—can still produce incorrect conclusions. AgriChat exhibits severe semantic drift and attributes the symptoms to ``Nectarine disease''. mRAG yields ``Peach anthracnose'', a related but inaccurate label, which indicates that noisy information from the retrieval process can mislead the model; VisRAG similarly gives the wrong answer ``Peach spot disease''. In contrast, Agri-MRAG once again demonstrates its superiority: it not only correctly identifies ``Peach scab'' but also gives a precise description of the red patch characteristics seen in the image, thereby confirming the effectiveness of our approach.

{Synthesizing the analysis of these two cases, we conclude that in the highly specialized vertical domain of agricultural plant pathology, general-purpose MLLMs face substantial challenges related to knowledge hallucination. Whether Qwen3-VL or the LLaVA series, these models are highly susceptible to misjudgment due to training data distribution biases when confronted with visually similar diseases or complex fruit symptoms. Conversely, retrieval-augmented frameworks such as Agri-MRAG exhibit greater robustness by integrating real-time retrieved agricultural knowledge with visual features.}

\begin{figure}[H]
\centering
\includegraphics[width=\textwidth]{case.pdf}
\caption{Examples are provided to illustrate the model responses on the PlantVillageVQA and PlantVQA test sets, where each case presents the input image alongside both the ground-truth (GT) label and the answer generated by the model. Correct predictions are highlighted in \textcolor{green}{green}, while incorrect findings are highlighted in \textcolor{red}{red}.\label{case}}
\end{figure}

\begin{figure}[H]
\centering
\includegraphics[width=\textwidth]{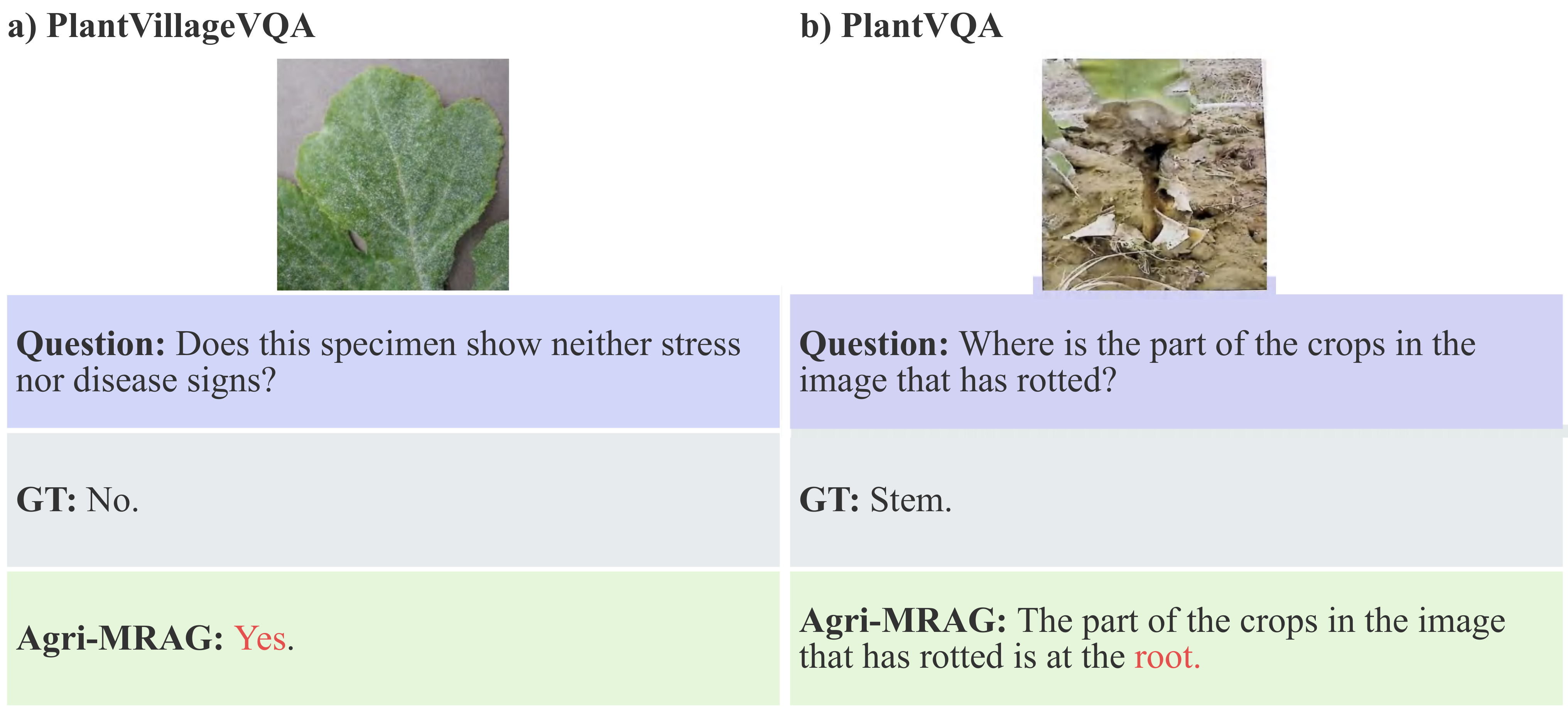}
\caption{{Representative failure examples of Agri-MRAG. The model produces incorrect predictions for fine-grained visual tasks, including plant stress and disease assessment and the localization of rotten crop tissues on the PlantVillageVQA and PlantVQA datasets.} \label{case2}}
\end{figure}

{To further analyze the performance of Agri-MRAG, we select one failure case each from PlantVillageVQA and PlantVQA, as illustrated in Figure~\ref{case2}. Agri-MRAG suffers from defects in fine‑grained visual localization and discrimination, which lead to misjudgments of crop disease status and rotting positions. Even with retrieval and parameterized knowledge incorporated, certain underlying visual encoding biases cannot be corrected. Moreover, the PPL-guided routing fails to detect such perceptual errors. These results indicate that Agri-MRAG requires further optimization in terms of local visual feature alignment and uncertainty recognition for perceptual errors.}

\subsection{Complexity Analysis of Parameterized Knowledge Injection}
First, we analyze the time complexity of the offline training phase. During training, the agricultural knowledge base consists of $ M $ image--text pairs, denoted as $ \mathcal{V} = \{(I_j, T_j)\}_{j=1}^{M} $. Each pair is augmented $ R $ times, yielding a total of $ MR $ training samples. For each sample, only the LoRA parameters $ A_j \in \mathbb{R}^{r \times a} $ and $ B_j \in \mathbb{R}^{b \times r} $ are optimized, where $ a $ and $ b $ denote the input and output dimensions of the FFN layer in the MLLM, respectively, and $ r \ll a, b $. In each training step, the dominant computation arises from the low-rank projection, which requires $ \mathcal{O}(r(a+b)) $. Therefore, the total training complexity over all samples is: $ \mathcal{O}_{train} = \mathcal{O}(MRr(a+b))$. By contrast, concatenation-based MRAG methods, such as mRAG and VisRAG, directly prepend $ k $ retrieved image--text pairs $ \{(I_j, T_j)\}_{j=1}^{k} $ to the input sequence for contextual enhancement and involve no parameter optimization, thus incurring no additional training cost. It should be noted that although the parameterized knowledge injection strategy introduces additional computational overhead during training, this cost is restricted to the offline phase. Since LoRA updates are independent across samples, the training process can be fully parallelized across data instances, further reducing the training time.

Second, we analyze the space complexity of the online inference phase. During inference, a subset of image--text pairs is first retrieved from the agricultural knowledge base $\mathcal{V}$, denoted as $C^{*}=\{(I_j,T_j)\}_{j=1}^{k}$. Each pair corresponds to a LoRA adapter $(A_j,B_j)$, and these adapters are fused via $\Delta \theta = \sum_{j=1}^{k} B_j A_j$. The computational complexity of fusing a single adapter is $\mathcal{O}(r(a+b))$. Therefore, the total complexity of the fusion stage is $\mathcal{O}\big(k r(a+b)\big)$. Subsequently, the fused model is forwarded through an $L$-layer transformer-based MLLM for inference. The standard transformer complexity per layer is $\mathcal{O}(s^2 d)$, where $d$ denotes the hidden size of the transformer backbone and $s$ denotes the sequence length. Furthermore, We define two inference routes under \S\ref{analysis:ppl}. In route 1, samples satisfying $\mathrm{PPL} < \tau$ do not introduce sequence expansion, and the sequence length is $s = n$, where $n$ denotes the input token length. Thus, the inference complexity is $\mathcal{O}_{route1} = L \cdot \mathcal{O}(n^2 d) + \mathcal{O}\big(k r(a+b)\big)$. In route 2, for concatenation-based MRAG, the input sequence is expanded to $n + km$, where $m$ denotes the average token length of each retrieved image--text pair. Therefore, its complexity becomes $\mathcal{O}_{MRAG} = L \cdot \mathcal{O}\big((n+km)^2 d\big)$. For samples satisfying $\mathrm{PPL} \geq \tau$, both the concatenation-based MRAG pathway  and the adapter fusion mechanism  are activated, yielding Equation~\eqref{analysis:eq1}.
\begin{equation}
\mathcal{O}_{route2} = \mathcal{O}_{MRAG} + \mathcal{O}_{route1} = L \cdot \mathcal{O}\big((n+km)^2 d\big) + \mathcal{O}\big(k r(a+b)\big)
\label{analysis:eq1}
\end{equation}

Let $p_{<\tau}$ and $p_{\geq\tau}$ denote the proportions of samples in the test set satisfying $\mathrm{PPL} < \tau$ and $\mathrm{PPL} \geq \tau$, respectively, with $p_{<\tau} + p_{\geq\tau} = 1$. Accordingly, the overall expected inference complexity is given by Equation~\eqref{analysis:eq2}.
\begin{equation}
\mathcal{O}_{infer} = p_{<\tau} \cdot \mathcal{O}_{route1} + p_{\geq\tau} \cdot \mathcal{O}_{route2} = L \cdot \mathcal{O}\big(p_{<\tau} n^2 d + p_{\geq\tau} (n+km)^2 d\big) + \mathcal{O}(k r(a+b))
\label{analysis:eq2}
\end{equation}

In our experiments, the proportions $p_{\geq\tau}$ in the PlantVillageVQA and PlantVQA datasets are 21.95\% and 14.11\%, respectively. Thus, under the realistic condition $p_{<\tau} > p_{\geq\tau}$, the proposed Agri-MRAG triggers the quadratic-complexity path only for a small fraction $p_{\geq\tau}$ of the samples, while the majority of the samples are processed through the more efficient route 1. In contrast, standard MRAG incurs a deterministic inference complexity of $L \cdot \mathcal{O}\bigl((n+km)^2 d\bigr)$ for every input. Consequently, in terms of average computational cost, the expected inference complexity of the proposed method is lower than that of standard.

Finally, we analyze the storage space complexity of the LoRA adapter. Let the parameter count of the MLLM be $\theta$, and let the storage cost per model parameter be $\gamma$ bytes, where $\gamma$ is used to uniformly represent the storage cost under different numerical precisions. Then, the space complexity of the MLLM can be expressed as $\mathcal{O}_{MLLM} = \mathcal{O}(\gamma \theta)$. For a LoRA adapter, let its low-rank decomposition structure be $A \in \mathbb{R}^{r \times a}$ and $B \in \mathbb{R}^{b \times r}$; then the parameter count of a single adapter is $\theta_{adapter} = r(a+b)$, and its corresponding space complexity is $\mathcal{O}(\gamma r(a+b))$. If there are $M$ adapters in the system, the overall space complexity is $\mathcal{O}_{adapter} = \mathcal{O}(\gamma M r(a+b))$. Therefore, the space complexity of the overall method can be uniformly expressed as $\mathcal{O}_{storage} = \mathcal{O}_{MLLM} + \mathcal{O}_{adapter} = \mathcal{O}\big(\gamma (\theta + M r(a+b))\big)$. In contrast, the standard MRAG method requires storing the original image-text pair data. Let the average storage overhead per image-text sample be $\delta$, then its storage complexity is $\mathcal{O}_{MRAG} = \mathcal{O}(\gamma \theta + \delta M)$.

{Taking Qwen3-VL-2B as an example, its storage size under FP32 precision is approximately $8.00\,\mathrm{GB}$.} When the LoRA rank is set to $r=2$, each LoRA adapter under FP32 occupies about $5.30\,\mathrm{MB}$ in storage. By comparison, a single image--text pair in the PlantVillageVQA and PlantVQA datasets accounts for merely $0.10\,\mathrm{MB}$ and $0.22\,\mathrm{MB}$ on average, respectively. This indicates that the storage overhead of a single LoRA adapter exceeds that of its corresponding original image--text pair. However, in real-world scenarios, data distributions often exhibit a long-tail characteristic, where the vast majority of samples are in a low-frequency or even zero-access state. For example, in the PlantVillageVQA dataset, image-text pairs with zero access account for approximately 85.16\%; pairs accessed 1 to 5 times account for about 8.08\%, and high-frequency pairs accessed more than 5 times account for only about 6.76\%. In the PlantVQA dataset, zero-access image-text pairs account for approximately 51.79\%; pairs accessed 1 to 5 times account for about 47.30\%, and high-frequency pairs accessed more than 5 times account for only about 0.91\%. Therefore, it is sufficient to construct adapters only for a small number of high-frequency and important agricultural image--text pairs to cover the majority of user queries, thereby reducing storage costs while maintaining performance.

{To quantify the actual deployment overhead, we report the resource consumption of trained and stored adapters in Table 10. Taking the current full training setting as an extreme example, we trained and stored 154,977 adapters for PlantVillageVQA and 12,038 adapters for PlantVQA. This massive number of adapters leads to a substantial disk usage of 801 GB and 63 GB, respectively. Meanwhile, using a single GPU, the total training time reaches 5.66 hours on PlantVillageVQA and 0.45 hours on PlantVQA, with a peak memory of approximately 4.3 GB. In terms of inference deployment, the loading and fusion latency is 0.29 s/adapter for PlantVillageVQA and 0.85 s/adapter for PlantVQA. The system adopts an incremental adapter generation and merging policy to support dynamic updates. These statistics reveal that although the per-adapter size is small, the cumulative storage and deployment overhead over massive datasets is non-negligible.}

{However, the specific experimental work on retaining only high-frequency samples remains to be carried out, and the storage overhead of adapters also remains a non-negligible issue. Therefore, we will conduct an in-depth study in future work to further address the long-tail phenomenon. Meanwhile, we will also devote efforts in future work to reducing the storage overhead of adapters.}

\begin{table}[H]
    
    \caption{{Statistics of trained and stored adapters and their deployment overhead on PlantVillageVQA and PlantVQA.}
    \label{adapter_stats}}
    \centering
    \resizebox{\textwidth}{!}{%
    \begin{tabular}{lcc}
        \toprule
        \textbf{Item} & \textbf{PlantVillageVQA} & \textbf{PlantVQA} \\
        \midrule
        Number of Trained Adapters & 154,977 & 12,038 \\
        Number of Stored Adapters & 154,977 & 12,038 \\
        Disk Usage & 801~GB & 63~GB \\
        Training Time (GPU-hours) & 5.66 & 0.45 \\
        Peak Memory & 4344~MB & 4346~MB \\
        Loading and Fusion Latency & 0.29 s/adapter & 0.85 s/adapter \\
        Update Policy & \makecell[l]{Incremental adapter\\generation and merging} & \makecell[l]{Incremental adapter\\generation and merging} \\
        \bottomrule
    \end{tabular}%
    }
\end{table}

\section{Discussion}

This study proposes the Agri-MRAG framework, which mitigates the critical challenges faced by MLLMs in agricultural vertical domains through three core mechanisms: semantic-consistency-guided evidence selection, parameterized knowledge injection, and PPL-guided uncertainty routing. The effectiveness of the framework can be deeply elucidated from the following three perspectives.

First, the semantic-consistency-guided evidence selection mechanism overcomes the limitation of a fixed retrieval quantity $k$. In fine-grained pest and disease diagnosis tasks, a predetermined $k$ easily leads to the omission of key pathological features or the influx of irrelevant noise, diverting the model's attention. By introducing cluster analysis in the multimodal embedding space, Agri-MRAG can automatically identify and exclude low-confidence outlier samples according to the degree of semantic alignment between the query and candidate evidence, retaining only highly cohesive and relevant knowledge subsets and thereby adaptively determining the optimal volume of evidence. This mechanism significantly improves the signal-to-noise ratio of the retrieved knowledge and strengthens the model's discriminative capability for agricultural disease data characterized by high inter-class similarity and significant intra-class variation. Ablation analysis demonstrates that the introduction of the clustering module serves as the primary driving force behind the performance leap, fundamentally suppressing noise interference and enabling the model to capture subtle pathological discriminative features more precisely when confronted with easily confusable visual symptoms.

Second, the parameterized knowledge injection strategy represents a fundamental shift in how external knowledge is integrated. Instead of following the common route of directly concatenating retrieved texts as explicit context, Agri-MRAG encodes the selected agricultural evidence into a set of low-rank adaptation parameters and merges these parameters into the model's feed-forward network via linear combination, thus achieving parameterized internalization of external knowledge. Such a design effectively avoids the attention dilution problem that arises from long contexts, letting the model tap into domain expertise in an implicit and tightly coupled fashion while keeping the backbone network weights frozen, hence adapting to specific agricultural diagnostic scenarios on the fly. This paradigm not only improves the semantic consistency of generated outputs but also grants the framework the ability to evolve continuously: when new pest or disease samples appear, one simply regenerates the adapter parameters and performs incremental merging, without retraining or fine-tuning the base model, thereby offering a practical path for timely agricultural knowledge updates. Ablation studies further show that although the adapter module may not deliver gains as large as those from the clustering module on certain metrics, its core value lies in the deep transformation it brings to knowledge fusion, making it an essential part of achieving trustworthy diagnosis.

Finally, the PPL-guided uncertainty routing mechanism, when deployed in practice, effectively balances the trade-off between inference efficiency and diagnostic accuracy. By quantifying the model's confidence when relying solely on internal parameterized knowledge, this mechanism adaptively selects an inference mode: for common pest and disease queries, the model exhibits low PPL and can respond rapidly based on implicit knowledge, significantly reducing computational latency; when confronted with complex symptoms, the model perceives higher uncertainty and triggers an explicit knowledge verification process, using fine-grained evidence to correct potential biases. {Unlike static models fine-tuned on large-scale domain data, Agri-MRAG dynamically determines how knowledge is used for each input sample based on its uncertainty, rather than always relying on a single fixed strategy.}

\section{Conclusion}
This paper presents Agri-MRAG, a multimodal retrieval-augmented framework for Agri-VQA that combines semantic-consistency-guided evidence selection, parameterized knowledge injection, and PPL-guided uncertainty routing. Agri-MRAG reduces retrieval noise by clustering retrieved image-text candidates and selecting high-confidence evidence subsets according to their relevance to the input image and question. It further encodes retrieved agricultural evidence into LoRA-based adapters, enabling external knowledge to guide response generation through parameter-level adaptation instead of relying only on prompt-level text concatenation. In addition, this study constructs PlantVQA, a region-specific Agri-VQA dataset for forestry and fruit crop pest and disease identification, containing 7,514 images and 15,028 expert-verified question-answer pairs. Experiments on PlantVillageVQA and PlantVQA show that Agri-MRAG outperforms general-purpose MLLMs, an agriculture-specific model named AgriChat, and the conventional mRAG and VisRAG baselines across BLEU, ROUGE-L, and METEOR metrics. In particular, Agri-MRAG achieves ROUGE-L scores of 41.21 and 46.58 on the two datasets, indicating its effectiveness in improving response quality and semantic consistency for fine-grained Agri-VQA.

Future work will focus on extending Agri-MRAG toward broader and more practical agricultural applications. The agricultural knowledge base and PlantVQA dataset will be further expanded to cover more crop species, disease categories, growth stages, and field conditions. Meanwhile we also plan to explore more efficient organization and deployment approaches for adapters so as to support scalable inference in real-world agricultural knowledge services; these efforts are expected to enhance the applicability and robustness of Agri-MRAG across a wider range of agricultural scenarios.

\vspace{6pt}

\authorcontributions{Conceptualization, X.Z. and F.L.; methodology, F.L.; software, F.L.; validation, X.L. and X.Z.; formal analysis, X.L.; investigation, X.Z.; resources, X.Z.; data curation, X.L.; writing---original draft preparation, F.L.; writing---review and editing, X.Z.; visualization, X.L.; supervision, X.Z.; project administration, X.Z.; funding acquisition, X.Z. All authors have read and agreed to the published version of the manuscript.}

\funding{This research was funded by the Central Guidance for Local Science and Technology Development Fund (Grant No. ZYYD2025QY19), the Bingtuan Science and Technology Program (Grant Nos. 2022DB005, 2018ZYYD-1), and the National Natural Science Foundation of China (NSFC) (Grant No. 62262057).}

\institutionalreview{Not applicable}

\dataavailability{This study utilized two datasets: PlantVillageVQA and PlantVQA. PlantVillageVQA is available through the HuggingFace (link: \url{https://huggingface.co/datasets/SyedNazmusSakib/PlantVillageVQA}). PlantVQA is a dataset independently constructed by this research team; please contact the corresponding author for access if needed.}

\acknowledgments{The authors would like to acknowledge the financial support provided by the Central Guidance for Local Science and Technology Development Fund (Grant No. ZYYD2025QY19) and the Bingtuan Science and Technology Program (Grant Nos. 2022DB005 and 2018ZYYD-1). The authors also thank all individuals and organizations who contributed to this study through technical assistance and valuable discussions.}

\conflictsofinterest{The authors declare no conflicts of interest.}

\begin{adjustwidth}{-\extralength}{0cm}

\reftitle{References}
\bibliography{ref_new}

\PublishersNote{}
\end{adjustwidth}
\end{document}